\documentclass[pdflatex,bst/sn-mathphys-num]{sn-jnl}

\usepackage{graphicx}%
\usepackage{multirow}%
\usepackage{amsmath,amssymb,amsfonts}%
\usepackage{amsthm}%
\usepackage{mathrsfs}%
\usepackage[title]{appendix}%
\usepackage{xcolor}%
\usepackage{textcomp}%
\usepackage{manyfoot}%
\usepackage{booktabs}%
\usepackage{algorithm}%
\usepackage{algorithmicx}%
\usepackage{algpseudocode}%
\usepackage{listings}%
\usepackage{overpic}
\usepackage{fvextra}
\DefineVerbatimEnvironment{PromptVerbatim}{Verbatim}
  {breaklines=true, breakanywhere=true}

\theoremstyle{thmstyleone}%
\theoremstyle{thmstyletwo}%

\theoremstyle{thmstylethree}%

\begin{document}

\title[Article Title]{Social Simulations with Large Language Model Risk Utopian Illusion}


\author[1]{\fnm{Ning} \sur{Bian}}\email{ningbian@cumt.edu.cn}

\author[2]{\fnm{Xianpei} \sur{Han}}\email{xianpei@iscas.ac.cn}

\author[2]{\fnm{Hongyu} \sur{Lin}}\email{hongyu@iscas.ac.cn}

\author[3]{\fnm{Baolei} \sur{Wu}}\email{blwu@cumt.edu.cn}

\author*[1]{\fnm{Jun} \sur{Wang}}\email{wj999lx@cumt.edu.cn}

\affil[1]{\orgdiv{School of Information and Control Engineering}, \orgname{China University of Mining and Technology}, \orgaddress{\city{Xuzhou}, \postcode{221116}, \state{Jiangsu}, \country{China}}}

\affil[2]{\orgdiv{Chinese Information Processing Laboratory}, \orgname{Institute of Software, Chinese Academy of Sciences}, \orgaddress{\city{Beijing}, \postcode{100190}, \country{China}}}

\affil[3]{\orgdiv{School of Computer Science and Technology}, \orgname{China University of Mining and Technology}, \orgaddress{\city{Xuzhou}, \postcode{221116}, \state{Jiangsu}, \country{China}}}


\abstract{Reliable simulation of human behavior is essential for explaining, predicting, and intervening in our society.
Recent advances in large language models (LLMs) have shown promise in emulating human behaviors, interactions, and decision-making, offering a powerful new lens for social science studies.
However, the extent to which LLMs diverge from authentic human behavior in social contexts remains underexplored, posing risks of misinterpretation in scientific studies and unintended consequences in real-world applications.
Here, we introduce a systematic framework for analyzing LLMs’ behavior in social simulation. Our approach simulates multi-agent interactions through chatroom-style conversations and analyzes them across five linguistic dimensions, providing a simple yet effective method to examine emergent social cognitive biases. We conduct extensive experiments involving eight representative LLMs across three families.
Our findings reveal that LLMs do not faithfully reproduce genuine human behavior but instead reflect overly idealized versions of it, shaped by the social desirability bias. In particular, LLMs show social role bias, primacy effect, and positivity bias, resulting in ``Utopian'' societies that lack the complexity and variability of real human interactions.
These findings call for more socially grounded LLMs that capture the diversity of human social behavior.
}





\maketitle

\section{Introduction}

Modeling human society is challenging because behavior emerges from the complex interplay of individual cognition, social interactions, and environmental constraints. While traditional observational and experimental approaches provide valuable insights, they are often limited in scale, scope, and ethical feasibility \cite{jimenez2021experimental}.
Systems that can accurately simulate human behavior and cognition are therefore becoming indispensable tools for social science \cite{gilbert2000build, davidsson2002agent}. They help us explain, predict, and intervene in social dynamics in ways that are efficient, affordable, and safe. For example, they can help anticipate panic buying during crises like the COVID-19 pandemic \cite{fan2024agent, fu2021propagation}, enhance coordination of emergency responses during disasters \cite{pan2007multi}, and mitigate the spread of misinformation on social media \cite{butler2024mis, tambuscio2015fact}.

Computational simulations provide a powerful foundation for these advances, enabling researchers to model, test, and refine theories in social science in controlled, repeatable, and scalable ways \cite{hofman2021integrating}. By enabling detailed analysis of hypothetical scenarios and counterfactuals, these methods can reveal insights that are difficult or impossible to obtain through direct empirical observation, opening new avenues for both scientific understanding and policy intervention. Conventional social simulation systems build on centuries of accumulated social and behavioral knowledge to create agents \cite{conte2014agent, chandramouli2024workflow}. However, these models face limitations: they often require extensive domain expertise to design, are computationally intensive, and can be slow to adapt to new social phenomena. 

In recent years, large language models (LLMs) such as ChatGPT \cite{chatgpt}, LLaMA \cite{touvron2023llama}, and DeekSeek~\cite{guo2025deepseek} have demonstrated emerging abilities in replicating human behavior by learning directly from human-generated data \cite{strachan2024testing, li2023camel, kosinski2024evaluating, piao2025agentsociety, dillion2023can, griffin2023susceptibility, argyle2023out, aher2023using, park2022social}.
For example, LLMs exhibit human-like intuitive behavior and reasoning biases in semantic illusion tests, mirroring human's System 1 thinking \cite{hagendorff2023human}. They have also demonstrated theory-of-mind reasoning ability, performing at or above human levels on tasks involving false beliefs and indirect requests \cite{strachan2024testing}. Moreover, cognitive dissonance, a hallmark of human self-referential processing, emerged clearly in GPT‑4o \cite{lehr2025kernels}.
As a result, these models are being adopted as ``promising'' \cite{piao2025agentsociety} tools in computational social science to create human-like agents \cite{zhou2023sotopia, wang2024sotopia} and to analyze multi-agent interactions in complex social systems \cite{tang2024gensim, mou2024individual, gao2024large, yang2024oasis, park2023generative}. 

On the other hand, accumulating evidence also suggests that LLMs exhibit systematic biases, both socially \cite{gallegos2024bias, kotek2023gender, navigli2023biases, plaza2024angry, demidova2024john, wan2024white, plaza2024divine} and cognitively \cite{bian2024influence, koo2024benchmarking, itzhak2024instructed, chen2025manager, hu2025generative}. For example, Hewitt et al. \cite{hewitt2024predicting} showed that LLMs introduce systematic errors when predicting social science outcomes. Biases such as sycophancy \cite{cfeae38342824620aafa35192bd0fda8} and emotional positivity \cite{bian2024influence} are reinforced through preference alignment \cite{ouyang2022training}. LLMs also exhibit non-humanlike reasoning under uncertainty at theory-of-mind tasks \cite{strachan2024testing}. Because LLMs are typically optimized to be more helpful, honest, and harmless (3H) \cite{askell2021general, kamath2024tuning} than humans, their social behaviors may diverge considerably from ours \cite{fontana2025nicer}. Moreover, many social cognitive biases emerge only in multi-agent contexts \cite{ashery2025emergent}, making them difficult to detect through standard benchmark tests. 

This leads to a critical question: \textbf{To what extent can LLMs authentically replicate human social behavior in computational simulations?}
Answering this question is critical because, without understanding how LLMs deviate from real humans, researchers may draw incorrect conclusions about our society, and deployed systems could act in biased ways, leading to real-world consequences. Since LLMs can amplify social and cognitive biases, unexamined reliance on them risks misleading scientific insights and undermining decision-making \cite{demszky2023using}. Furthermore, as AI becomes increasingly embedded in our social systems, it is important to ensure that agents can interact effectively with the complex, variable, and unpredictable behavior of humans \cite{bail2024can}. However, most existing studies focus on single-agent benchmarks, leaving a critical gap in our understanding of how LLMs behave in dynamic multi-agent social contexts. In this study, we define authentic human behavior as social interaction patterns that approximate the statistical and psychological properties of real human communication, including diversity of roles and language use, emotional variability, and adaptive topic shifts. 

In this paper, we address this question by focusing on three key aspects:

1) How do LLM-simulated agents differ from humans in social interaction patterns?

2) How do social cognitive biases, especially those that emerge in collective contexts, develop through interactions among LLM agents?

3) What are the underlying sources of these biases in LLMs?


\begin{figure*}[!t]
  \centering  
  \includegraphics[width=\columnwidth]{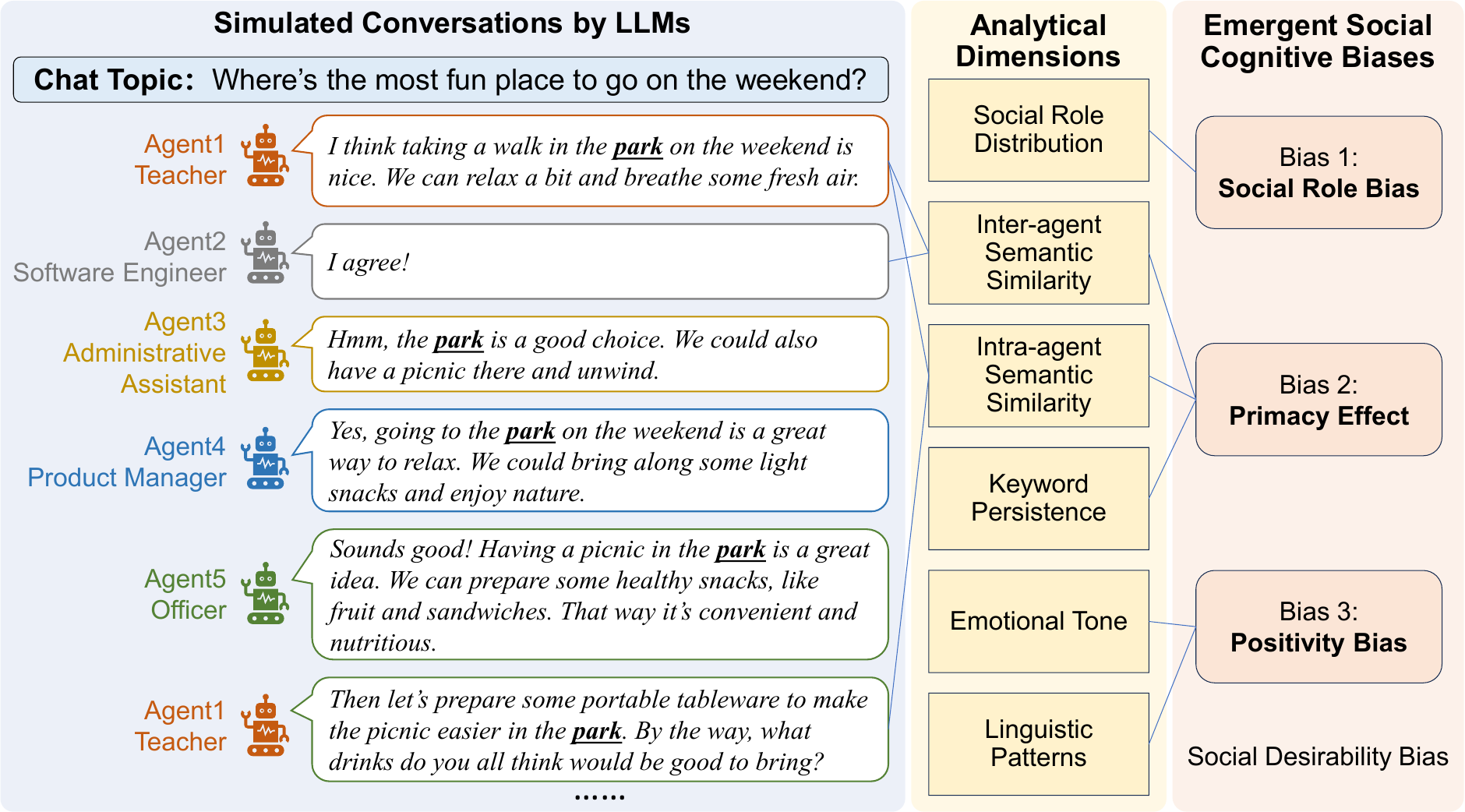}  
  \caption{\textbf{Framework for evaluating authenticity of LLM-driven social simulations.}
Role-conditioned agents converse in multi-agent chatrooms; dialogues are analyzed on five dimensions, including role distribution, semantic similarity, keyword persistence, sentiment, and linguistic features to identify systematic divergences from human interaction.}
  \label{figure-f0}
\end{figure*}

We present a systematic framework to analyze LLMs in social simulations, as shown in Fig. \ref{figure-f0}. We designed a simple yet effective method for analyzing multi-agent interactions through chatroom-style conversations on given topics, enabling the identification of social cognitive biases that emerge in group dynamics. To emulate the diversity of agents, each agent was assigned a unique social role reflecting the behaviors and responsibilities expected of individuals in society \cite{eagly2012social, chen2024persona}. LLMs were instructed to generate utterances consistent with both their role and the ongoing conversation. We conducted experiments on 8 representative LLMs from 3 families (GPT, LLaMA, and DeepSeek). To assess the authenticity of these interactions, we analyzed LLM-generated chats against human conversations across five dimensions covering social role distribution, semantic similarity, keyword persistence, sentiment, and linguistic pattern. 
To investigate the origins of biases in LLM-driven social simulation, we further analyzed both large-scale textual corpora and preference datasets. We examined the frequency of occupational terms in the Google Books Ngram Dataset v3 \cite{michel2011quantitative} to approximate how social roles are represented in the textual corpora that contribute to LLM training. We also analyzed sentiment and semantic alignment in widely used preference datasets, including PRISM \cite{kirk2024prism} and UltraFeedback \cite{cui2024ultrafeedback}, to assess how reinforcement learning from human feedback may shape LLM outputs. Together, these analyses help explain how both pretraining data and alignment procedures contribute to biases in LLM-driven group interactions.

We report three key findings: (1) LLM-simulated agents diverge significantly from humans by producing polite, homogeneous, and overly coherent dialogues. Their conversations show inflated semantic similarity between utterances, reduced diversity of roles, and diminished linguistic and emotional variability, yielding a ``Utopian'' but unrealistic form of social interaction. (2) LLMs display three social cognitive biases in a multi-agent setting: social role bias (overrepresentation of high-status, idealized professions), primacy effect (early-introduced topics dominate discourse), and positivity bias (persistent preference for agreeable and optimistic tone). (3) Textual datasets overrepresent prestigious roles, and preference datasets favor emotionally positive and semantically redundant responses, both contributing to these biases. 

\section{Results}
\subsection{Generated social roles}

To analyze how LLMs represent social roles, we first examined the distribution of occupations in the social roles generated by LLMs and compared it to global occupation distributions. To systematically categorize the generated roles, we mapped each role onto the \textit{International Standard Classification of Occupations 2008 (ISCO-08)}\footnote{\url{https://isco.ilo.org/}}, a widely used framework developed by the International Labour Organization (ILO). Using a reasoning-oriented LLM, we automatically classified each role into one of nine ISCO-08 occupational categories based on detailed class descriptions. The automated classification was further validated through manual annotation (see Methods).

\begin{figure*}[!t]
  \centering  
  \includegraphics[width=\columnwidth]{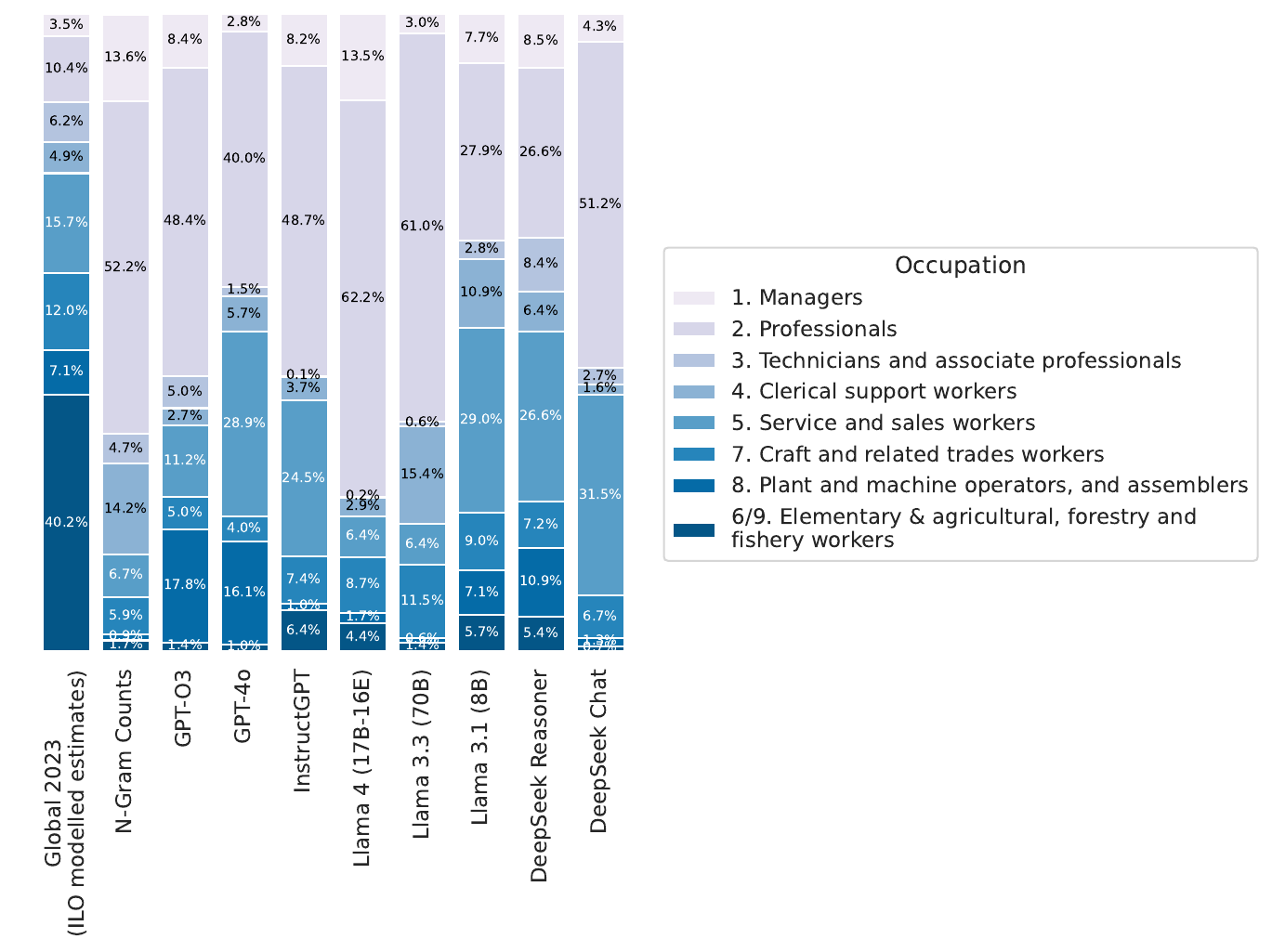}  
  \caption{Occupational distribution of social roles generated by LLMs compared to global employment estimates by ILO and n-gram counts of occupation examples in the Google Books Ngram Dataset v3. $N=5,300$ for Llama 3.1 (8B) and $N=1,100$ for other models. }
  \label{figure-f1}
\end{figure*}

As shown in Fig.~\ref{figure-f1}, the distribution of occupations generated by the LLMs differs significantly from the ILO's modelled global employment shares in 2023\footnote{\url{https://www.ilo.org/publications/flagship-reports/world-employment-and-social-outlook-may-2025-update}}. 
In human societies, the majority of individuals work in labor-intensive and low-prestige occupations, including elementary and agricultural, forestry, and fishery work (Occupation 6/9, covering 40.2\%), forming the backbone of global economic activity. In contrast, this category is consistently under-represented in the LLM-generated roles, with rates as low as 0.7\% (DeepSeek Chat) to a maximum of only 6.4\% (InstructGPT).
Conversely, high-status occupations are heavily over-represented. Professionals (Occupation 2) account for only 10.4\% of the global workforce but dominate the generated roles, ranging from 26.6\% (DeepSeek Reasoner) to 62.2\% (Llama4). Similarly, managers (Occupation 1) also appear more often than their real-world share of 3.5\%, reaching up to 13.5\% (Llama4). Other categories, such as clerical support workers (Occupation 4), are over-represented in some models. For example, Occupation 4 accounts for 15.4\% of roles generated by Llama3.3, compared to 4.9\% globally.

To investigate the potential source of these biases, we analyzed the frequency of ISCO-08 occupation examples in the Google Books Ngram Dataset v3 \cite{michel2011quantitative}, shown as ``N-gram Counts'' in Fig.~\ref{figure-f1}. These frequencies serve as a proxy for how often various occupational terms appear in large-scale textual corpora, which form part of the training data for many LLMs. The distribution reveals a pronounced reporting bias: high-status occupations are vastly over-represented, while lower-prestige and labor-intensive roles are largely absent. Specifically, references to professionals (Occupation 2) account for 52.2\% of all occupation-related n-grams, and managers (Occupation 1) comprise 13.6\%, totaling over 65\% for just these two groups. In contrast, elementary occupations and agricultural, forestry, and fishery workers (Occupation 6/9) make up only 1.7\% of all occupation mentions, with similarly low rates for other manual labor categories. 

These findings suggest that LLMs systematically inherit reporting biases from their training corpora, giving rise to a ``social role bias''. Social roles that are more frequently mentioned in books and other textual sources, typically higher-status or aspirational roles such as software engineers, product managers, or school teachers, are more likely to be generated by the models. In contrast, common yet lower-prestige roles, such as elementary workers, farm laborers, or machine operators, are markedly under-represented. As a result, LLM-generated outputs present an overly idealized and selective view of social roles, diverging substantially from real-world distributions. This over-representation of ideal roles risks reinforcing social hierarchies and propagating an unrepresentative vision of society in social simulations.

To mitigate this distortion, we also included a set of human-authored personas from the PersonaChat dataset \cite{gopalakrishnan2023topical} besides the LLM-generated roles to create agents.

\subsection{Semantic similarity between utterances}
After generating the simulated conversations, we analyzed semantic similarity between consecutive utterances, both between agents (inter-agent) and within the same agent (intra-agent) in conversations. Cosine similarity was computed based on BERT-Large \cite{devlin-etal-2019-bert} embeddings.

\begin{figure*}[!t]
  \centering
  
  \begin{minipage}{0.49\linewidth}
    \centering
    \begin{overpic}[abs,width=\linewidth]{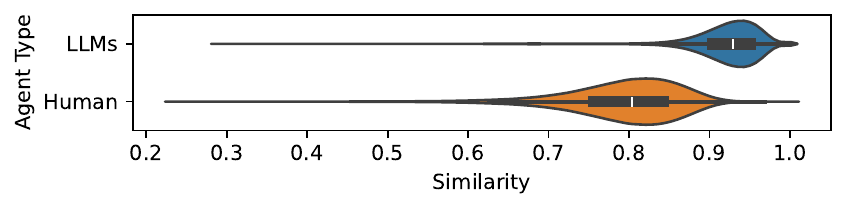}
      \put(3,45){{\small\fontfamily{phv}\selectfont\textbf{a}}}
    \end{overpic}
    \begin{overpic}[abs,width=\linewidth]{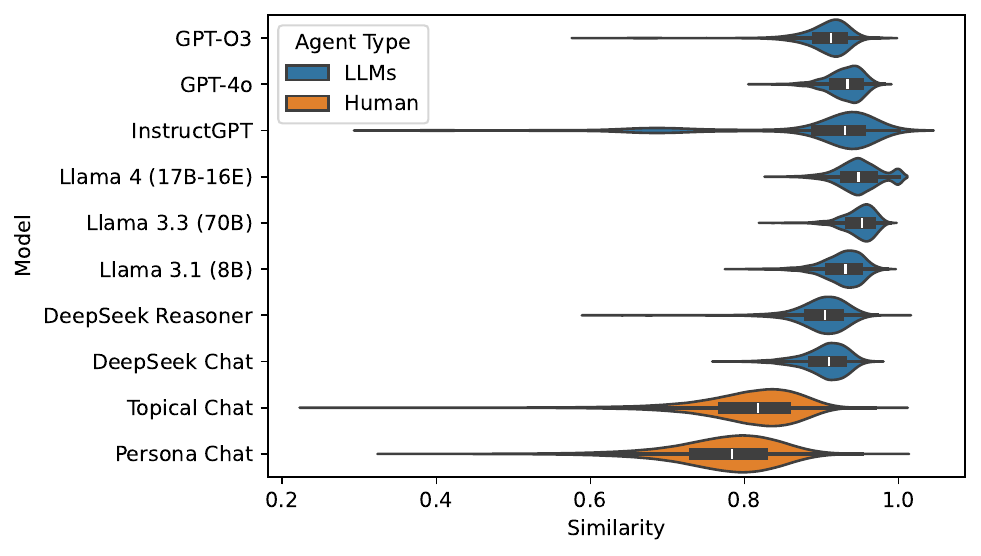}
      \put(3,95){{\small\fontfamily{phv}\selectfont\textbf{b}}}
    \end{overpic}
    \begin{overpic}[abs,width=\linewidth]{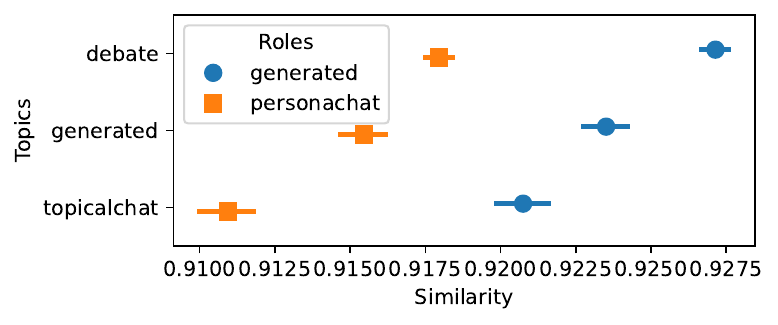}
      \put(3,65){{\small\fontfamily{phv}\selectfont\textbf{c}}}
    \end{overpic}
  \end{minipage}%
  \begin{minipage}{0.49\linewidth}
    \centering
    \begin{overpic}[abs,width=\linewidth]{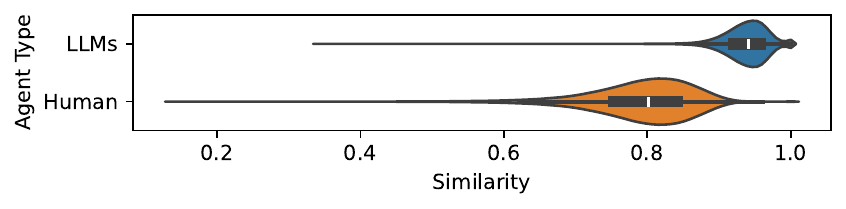}
      \put(3,45){{\small\fontfamily{phv}\selectfont\textbf{d}}}
    \end{overpic}
    \begin{overpic}[abs,width=\linewidth]{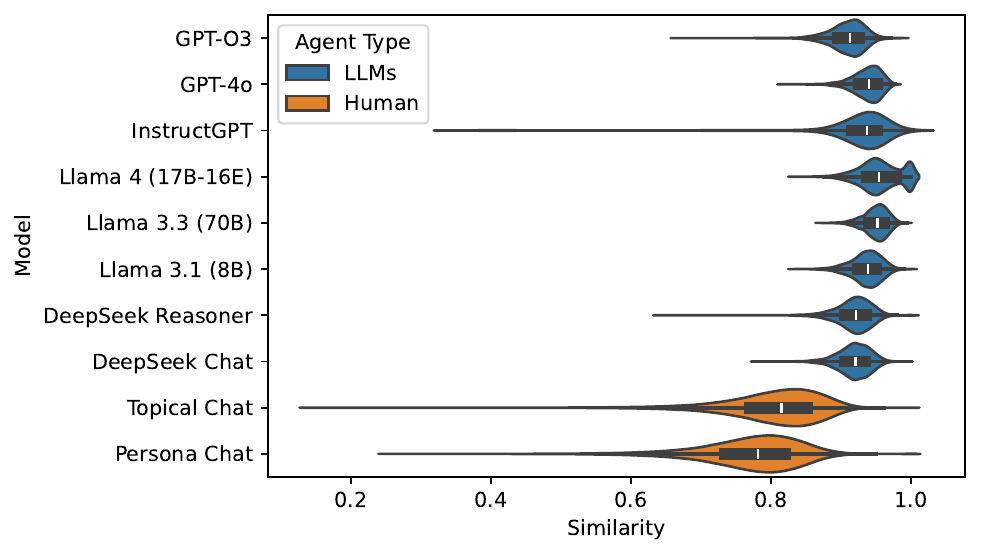}
      \put(3,95){{\small\fontfamily{phv}\selectfont\textbf{e}}}
    \end{overpic}
    \begin{overpic}[abs,width=\linewidth]{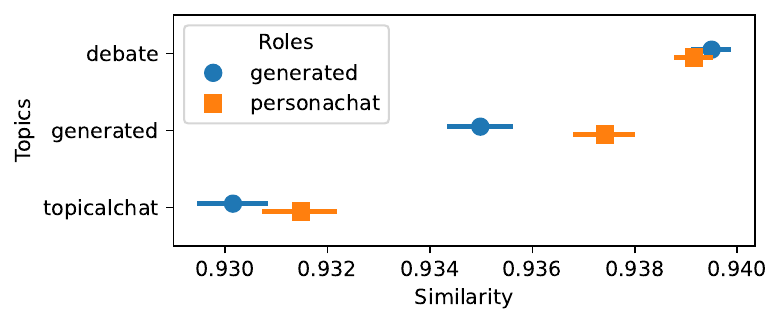}
      \put(3,65){{\small\fontfamily{phv}\selectfont\textbf{f}}}
    \end{overpic}
  \end{minipage}
  
  \caption{\textbf{Semantic similarity of chat utterances.}  
        \textbf{a} Semantic similarity between consecutive utterances in conversations generated by LLMs compared to human chat records from PersonaChat and TopicalChat 
        \textbf{b} Semantic similarity between consecutive utterances in conversations generated by different LLMs (with two chat participants) compared to human chat records from PersonaChat and TopicalChat. 
        \textbf{c} Semantic similarity between consecutive utterances in conversations conditioned on different sources of chat topics and social roles. $N=55,684$ for each point with debate topics and $N=18,846$ for each point with the other two topic sources. 
        \textbf{d} Semantic similarity between consecutive utterances generated by the same agent in LLM-generated and human conversations.  
        \textbf{e} Semantic similarity between consecutive utterances generated by the same agent across different LLMs and human conversations.  
        \textbf{f} Semantic similarity between consecutive utterances generated by the same agent conditioned on different sources of chat topics and social roles. $N=45,259$ for each point with debate topics and $N=15,259$ for each point with the other two topic sources. 
        All error bars represent 95\% confidence intervals.
    }

  \label{figure-f2}
\end{figure*}

\begin{figure*}[!t]
  \centering 
  \begin{minipage}{\linewidth}
    \centering
    \begin{overpic}[abs,width=\linewidth]{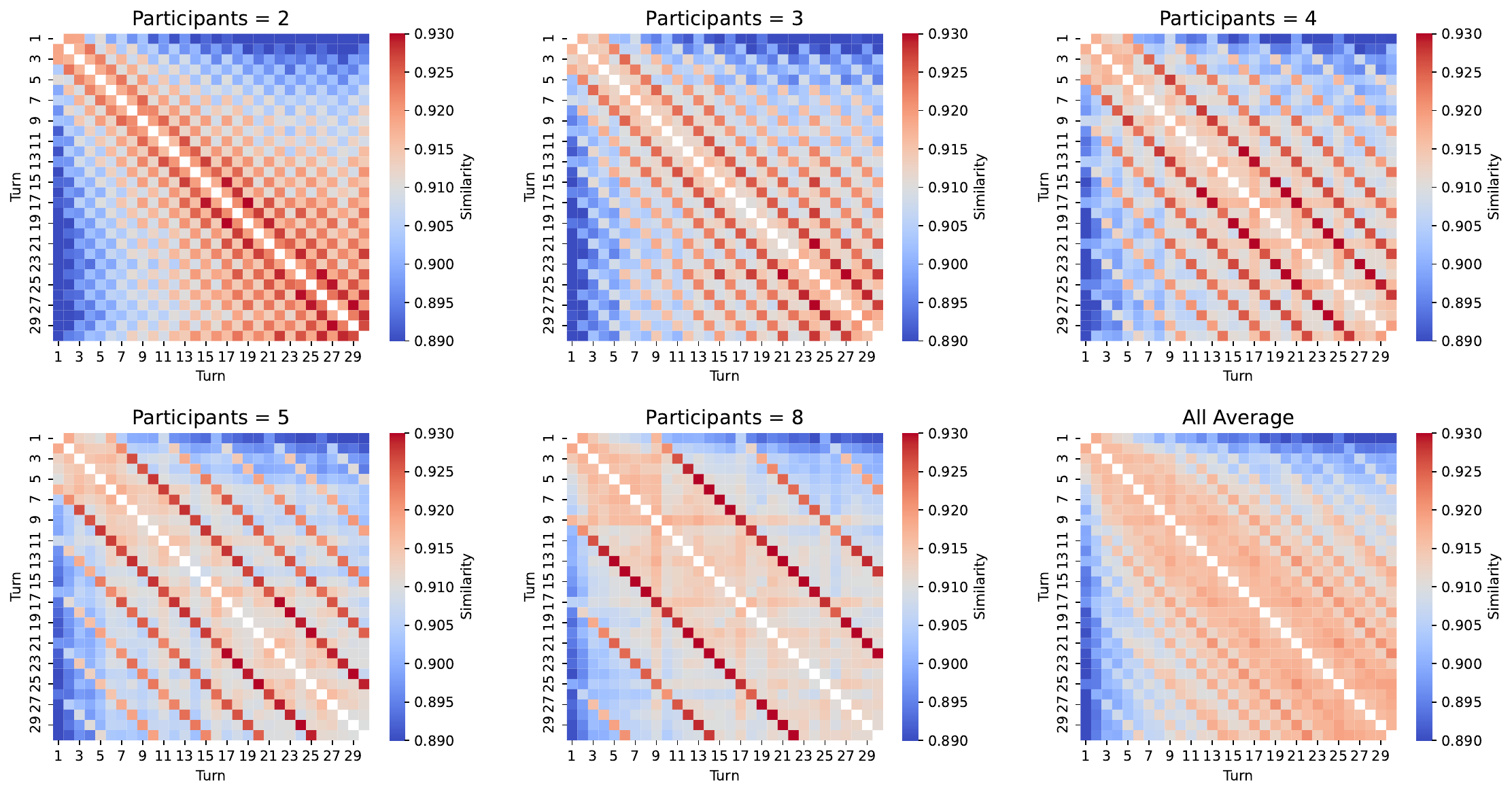}
    \end{overpic}
  \end{minipage}
  
  \caption{\textbf{Similarity matrices for the first 30 turns in simulated multi-agent dialogues.} Each heatmap shows average pairwise cosine similarity between utterances at different turns (x- and y-axes), grouped by number of participants (2, 3, 4, 5, 8) and overall average. Warmer colors (red) indicate higher semantic similarity. Diagonal (self-similarity) is masked.}

  \label{figure-f3}
\end{figure*}

We found that LLM-generated conversations exhibit significantly higher semantic similarity both between agents and within the same agent's turns compared to human group chats (Fig. \ref{figure-f2}a, $p<0.0001$, and Fig. \ref{figure-f2}d, $p<0.0001$). This suggests that LLMs tend to produce more semantically redundant utterances. This pattern holds consistently across all tested models (Fig. \ref{figure-f2}b and Fig. \ref{figure-f2}e), with human chat records exhibiting greater variation and lower similarity ($p<0.0001$ for all comparisons). Among the models, the reasoning-oriented GPT-O3 and the two DeepSeek models show slightly lower inter-agent and intra-agent similarities than the other models ($p<0.0001$ compared with the overall average), though still higher than humans, indicating somewhat more diverse outputs.
These results indicate that LLM-generated conversations show more semantic repetition, whereas human conversations show more diversity and semantic distance between utterances.

The analyses in Fig. \ref{figure-f2}c and \ref{figure-f2}f illustrate how topic and social role sources influence utterance similarity. Chat topics were drawn from three complementary sources to ensure diversity and representativeness: (i) randomly generated by the LLM itself, (ii) keyword combinations sampled from the TopicalChat dataset \cite{gopalakrishnan2023topical}, and (iii) debate topics used in prior studies \cite{salvi2025conversational}. In both inter-agent (Fig. \ref{figure-f2}c) and intra-agent (Fig. \ref{figure-f2}f) settings, conversations on the debate topic exhibit the highest semantic similarity, while those based on TopicalChat topics show the lowest. This suggests that debates tend to elicit more repetitive and narrowly focused exchanges, whereas open-ended, human-authored topics foster greater diversity. Furthermore, Fig. \ref{figure-f2}c shows that when social roles are generated by LLMs, inter-agent utterance similarity is higher than when roles are from PersonaChat, implying that authentic, human-written roles encourage more varied and less redundant dialogue. In contrast, Fig. \ref{figure-f2}f reveals the opposite trend for intra-agent similarity. Conversations with PersonaChat roles often exhibit higher similarity, except in the debate topic, where no significant difference is observed. This pattern suggests that fixed, human-authored personas may lead individual agents to reuse consistent linguistic and thematic patterns across their turns, increasing within-agent redundancy, while LLM-generated roles encourage more flexible self-expression. 

We further constructed similarity matrices of the first 30 turns of dialogues, as shown in Fig. \ref{figure-f3}. Each heatmap depicts the pairwise semantic similarity between turns with 2, 3, 4, 5, or 8 participants, as well as the overall average. A clear banded structure emerges, reflecting the periodic recurrence of the same agent's turns. Since individual agents tend to maintain consistent perspectives and language across their own turns, utterances by the same agent exhibit higher semantic similarity, producing parallel bands above the diagonal at intervals corresponding to the number of participants. 
As the dialogue progresses from turn 1 to 30, we also observe a drift toward higher similarity, indicated by warmer (red) colors and fewer cooler (blue) regions in later columns of the heatmaps. This pattern suggests that conversations gradually become more repetitive and filled with empty verbiage as agents converge on shared topics or viewpoints.

These linguistic patterns have broader implications for how LLMs diverge from human conversational patterns. Human conversations are inherently dynamic, reflecting negotiation, conflict, and diversity of perspective. The higher semantic redundancy observed in LLM-generated dialogues produces an artificially uniform discourse, lacking the argumentative and divergent tendencies typical of real social interactions. As a result, such simulations risk underestimating the diversity and adaptability of real social groups, potentially limiting their utility for understanding collective decision-making and social influence processes.

\subsection{Keyword Persistence}

\begin{figure*}[!t]
  \centering
  
  \begin{minipage}{0.49\linewidth}
    \centering
    \begin{overpic}[abs,width=\linewidth]{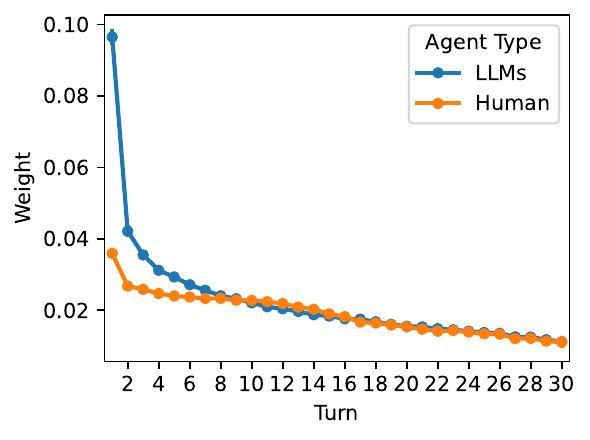}
      \put(0,130){{\small\fontfamily{phv}\selectfont\textbf{a}}}
    \end{overpic}
  \end{minipage}%
  \begin{minipage}{0.49\linewidth}
    \centering
    \begin{overpic}[abs,width=\linewidth]{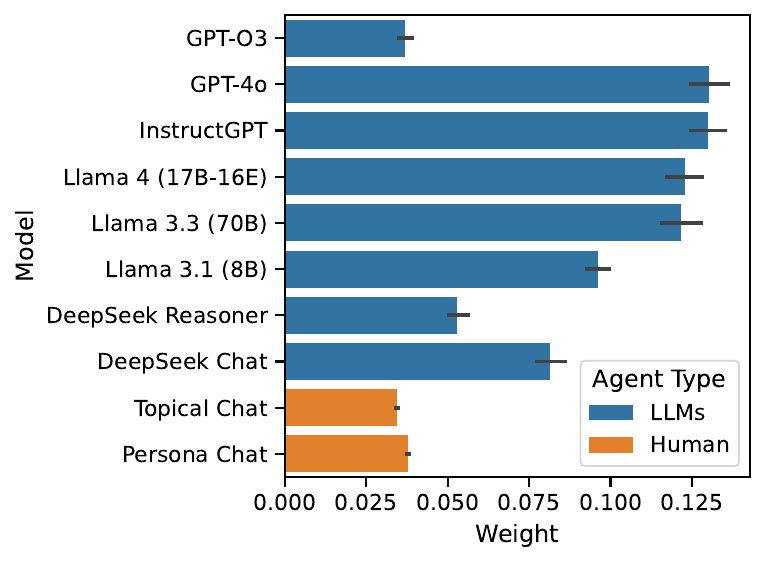}
      \put(5,130){{\small\fontfamily{phv}\selectfont\textbf{b}}}
    \end{overpic}
  \end{minipage}
  
    \caption{\textbf{Keyword weights and primacy.}
        \textbf{a} Average weight of keywords as a function of the turn in which a keyword first appears, comparing LLM-generated and human conversations. 
        \textbf{b} Average weight of keywords introduced in the first utterance, by model family and humans. All error bars represent 95\% confidence intervals.
    }
  \label{figure-keywords}
\end{figure*}

To explore how topics evolve throughout a conversation, we examined the persistence of keywords across conversation turns. For each utterance, we extracted keywords and their importance weights using KeyBERT \cite{grootendorst2020keybert} and tracked their influence throughout the conversation. For each keyword, we recorded the turn in which it first appeared and computed its average importance across all turns, assigning a weight of 0 for turns where the keyword did not appear. A higher average weight indicates that keywords introduced at that point continued to have a strong influence on subsequent conversations, reflecting greater topical persistence.

Fig.~\ref{figure-keywords}a shows the average keyword weights as a function of the turn in which each keyword first appears. Compared to human conversations, LLM-generated dialogues exhibit stronger persistence of early keywords (within the first seven turns). In other words, keywords introduced earlier tend to retain higher weights throughout the dialogue, reflecting a pronounced \textit{primacy effect} \cite{murdock1962serial} in LLM-driven conversations. This suggests that once a topic is introduced, it disproportionately shapes subsequent content, serving as a semantic anchor. For example, as shown in Fig.~\ref{figure-f0}, once the keyword ``park'' appears in the first turn, it continues to dominate subsequent turns. Each agent repeatedly refers to the ``park'' and elaborates on related activities such as picnicking and relaxing, rather than introducing new locations or shifting the topic. This illustrates how early keywords strongly guide the direction of LLM-generated dialogue, leading to high topical persistence.

Fig.~\ref{figure-keywords}b further examines keywords introduced in the first turn. In most LLM-generated conversations, these initial keywords have significantly higher average weights than those in human dialogues. An exception occurs with the two reasoning-oriented LLMs, GPT-o3 and DeepSeek-Reasoner, which exhibit lower initial keyword weights compared to other models. This suggests that reasoning models may permit greater topic flexibility than instruction-tuned models.

Taken together, these findings highlight the strong influence of topics introduced at the start of LLM-generated dialogues. Early topics tend to dominate the conversation, which helps maintain thematic coherence but also limits the natural evolution of ideas. In contrast, human conversations develop more flexibly, with themes shifting through negotiation, attention, and mutual adjustment. The primacy effect in LLM dialogues reduces this flexibility, producing discussions that are less reflective of collective sense-making among humans. As a result, social simulations based on LLMs may underestimate the adaptive and dynamic nature of human group interaction.

\subsection{Utterance sentiment}

\begin{figure*}[!t]
  \centering
  
  \begin{minipage}{0.49\linewidth}
    \centering
    \begin{overpic}[abs,width=\linewidth]{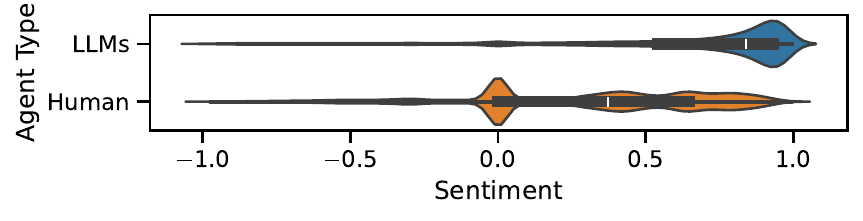}
      \put(0,50){{\small\fontfamily{phv}\selectfont\textbf{a}}}
    \end{overpic}
    \begin{overpic}[abs,width=\linewidth]{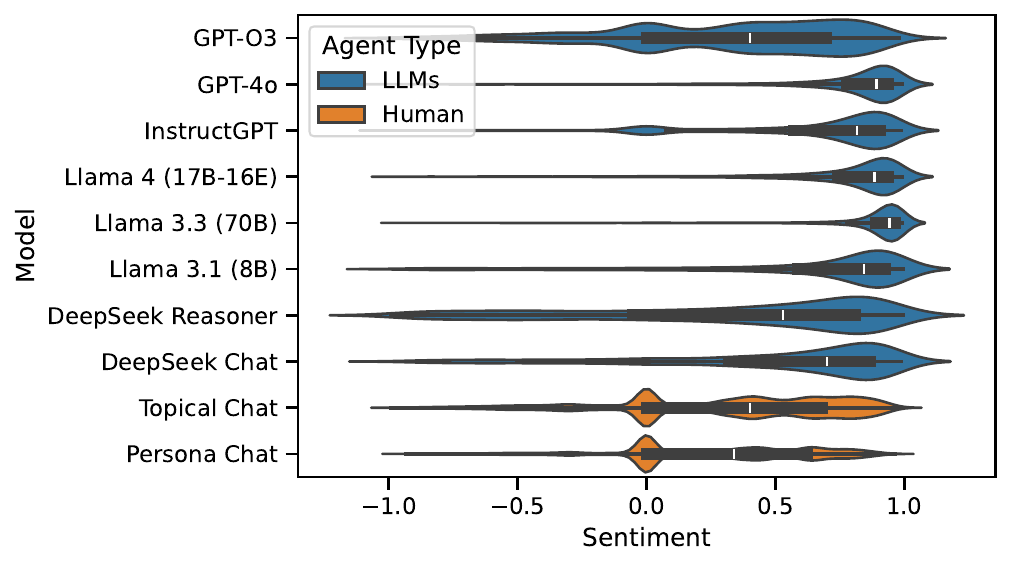}
      \put(0,100){{\small\fontfamily{phv}\selectfont\textbf{b}}}
    \end{overpic}
  \end{minipage}%
  \begin{minipage}{0.49\linewidth}
    \centering
    \begin{overpic}[abs,width=\linewidth]{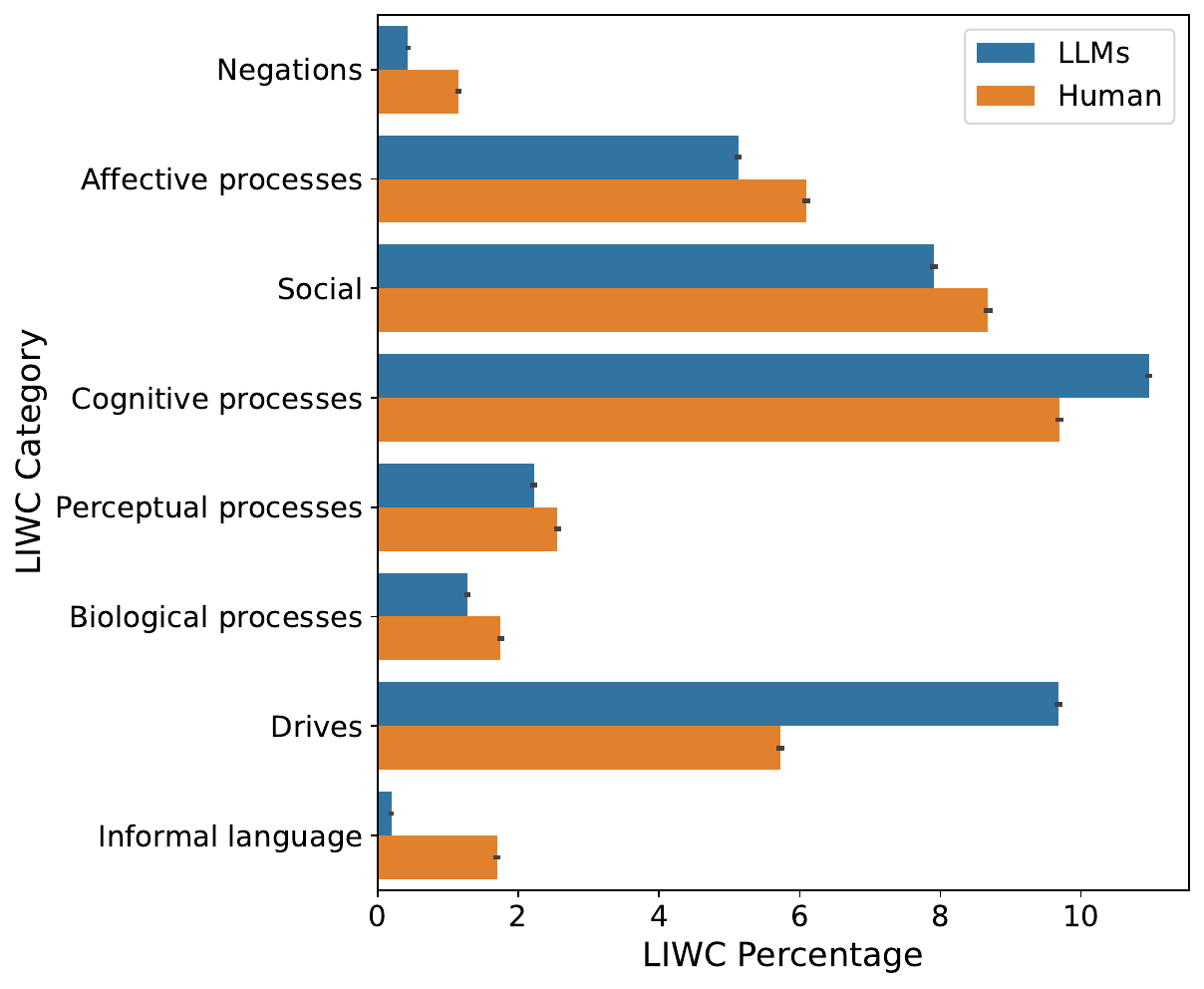}
      \put(1,150){{\small\fontfamily{phv}\selectfont\textbf{c}}}
    \end{overpic}
  \end{minipage}
  
  \caption{\textbf{Sentiment and linguistic patterns of simulated chat utterances.}
        \textbf{a} Sentiment distribution of utterances generated by LLMs compared to human chat records from PersonaChat and TopicalChat. 
        \textbf{b} Sentiment distribution of utterances generated by LLMs (with two chat participants) compared to human chat records from PersonaChat and TopicalChat. 
        \textbf{c} Linguistic patterns of utterances from LLMs and humans, analyzed using LIWC. All error bars represent 95\% confidence intervals. $N=178{,}824$ for LLMs and $N=319{,}816$ for humans.
    }
  \label{figure-f4}
\end{figure*}

We assessed the sentiment of utterances using VADER \cite{ahuja2017clustering}. We found that the sentiment score of conversations generated by LLMs was significantly more positive than that observed in human chat data (Fig.~\ref{figure-f4}a, $p < 0.0001$).
We further compared the sentiment across different LLMs, as shown in Fig. \ref{figure-f4}b. Overall, utterances generated by LLMs were significantly more positive than those in human chat records ($p<0.0001$), except for GPT-O3 on TopicalChat, where no significant difference was observed ($p=0.971$). Notably, the two reasoning models, GPT-O3 and DeepSeek-Reasoner, exhibited significantly lower sentiment scores than the other, non-reasoning LLMs ($p<0.0001$). Among them, GPT-O3 displayed a sentiment distribution most similar to that of the human chat datasets. These findings suggest that reasoning-oriented LLMs may generate more human-like emotional patterns than non-reasoning models. 

This tendency shows that LLMs produce consistently positive and agreeable responses during interactions. In multi-agent simulations, this positivity bias can create a reinforcing cycle, where agents amplify each other’s optimistic tone. While this makes conversations appear more harmonious, it reduces emotional diversity and realism compared to human interactions, which typically include a mix of positive, neutral, and negative emotions that reflect conflict, disagreement, and nuance. The resulting over-positivity in LLM-generated dialogues may mask social tensions and produce an overly harmonious yet unrealistic representation of group interaction, limiting the ability of such simulations to capture the role of emotional conflict in shaping real social dynamics.

\subsection{Linguistic patterns}
We compared the linguistic characteristics of LLM-generated and human-written utterances using LIWC \cite{tausczik2010psychological, pennebaker2015development}. As shown in Fig.~\ref{figure-f4}c, distinct patterns emerged between the two groups across several linguistic categories.

LLMs used significantly fewer disagreement and negation terms than humans ($p<0.0001$), indicating that expressions of dissent, criticism, or conflict were less frequent in simulated conversations. This pattern reflects an aversion to conflict in LLM-generated dialogue, contrasting with the more dynamic and argumentative nature of human group interactions.

LLMs also produced a higher proportion of words related to cognitive processes and motivational drives. Specifically, cognitive-process terms (e.g., think, know, because) appeared more frequently in LLM outputs, suggesting a stronger emphasis on reasoning and goal-directed communication. Similarly, drive-related language (e.g., words associated with achievement, power, and reward) was more prevalent in LLM-generated utterances, reflecting an increased focus on purpose and control.

In contrast, humans used significantly more social-process words, reflecting greater interpersonal engagement and attention to others during conversation. They also employed more affective-process terms and informal language, demonstrating richer emotional nuance and conversational informality. Additionally, humans used slightly more perceptual and biological process words, suggesting stronger grounding in sensory and bodily experience.

Sociologically, the predominance of cognitive and goal-directed language in LLM dialogues suggests a focus on reasoning at the expense of social and emotional processes that shape human interactions. Real human conversations rely heavily on interpersonal cues, emotional nuance, and informal communication to negotiate meaning, build trust, and maintain social cohesion. The reduced social and emotional richness in LLM-generated dialogues may therefore limit the ability of simulations to capture authentic patterns of collective coordination, persuasion, and relational dynamics.

\subsection{Scaling to larger chat groups}

To investigate how conversational patterns change with increasing group size while keeping computational costs under control, we simulated chatrooms with 2, 3, 4, 5, 8, 12, 16, 24, and 32 participants using the LLaMA 3.1 (8B) model. We measured semantic similarities and sentiment scores across these group sizes to assess how coherence and emotional tone evolve in larger simulated social settings.

\begin{figure*}[!t]
  \centering  
  \includegraphics[width=0.8\columnwidth]{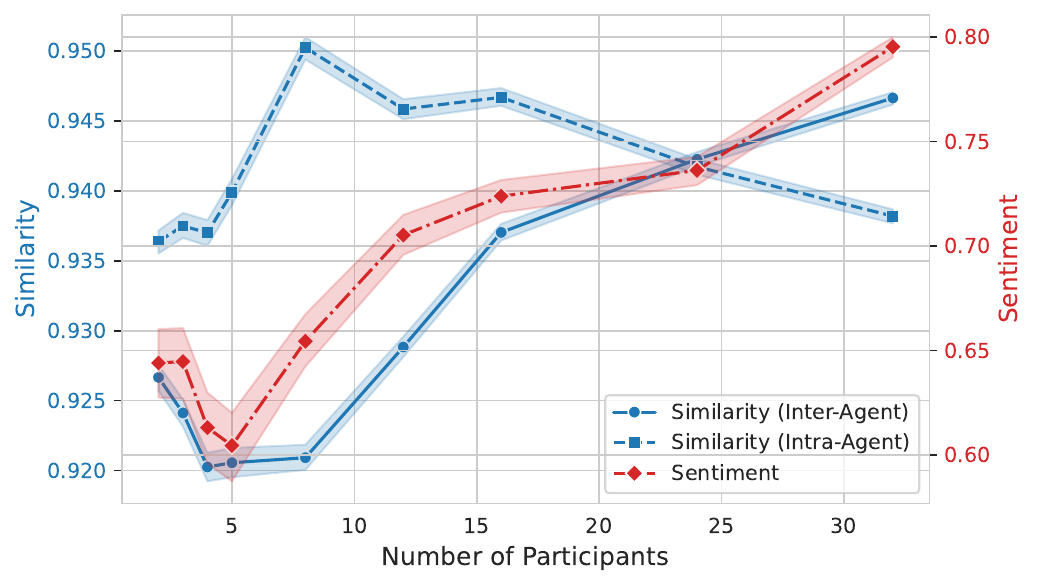}  
  \caption{\textbf{Effects of group size on semantic similarity and sentiment in simulated multi-agent conversations.} 
  The blue solid line represents inter-agent similarity (left axis), the blue dashed line represents intra-agent similarity (left axis), and the red line (right axis) indicates sentiment. Error bands denote 95\% confidence intervals.}
  \label{figure-f5}
\end{figure*}

As shown in Fig.~\ref{figure-f5}, inter-agent similarity initially decreases as group size increases from 2 to 4 participants. Beyond this point, similarity begins to rise, suggesting growing convergence or repetition in dialogue as groups become larger. This pattern may reflect a tendency toward uniformity or echoing behavior in large-group discussions. In contrast, intra-agent similarity peaks around 8 participants and gradually declines thereafter, likely because as the number of participants grows, each agent struggles to maintain coherence with its own prior contributions within an increasingly complex context.

Meanwhile, sentiment shows a consistent upward trend for group sizes above 5, indicating that the overall emotional tone becomes increasingly positive in larger groups. One possible explanation is that with more agents interacting, there emerges a collective bias toward politeness, affirmation, or sentiment smoothing.

These results indicate that increasing group size systematically reshapes LLM-generated conversations, producing greater linguistic uniformity and a more positive emotional tone. Unlike real social groups, where larger size often brings greater diversity of viewpoints, LLM simulations display a convergence toward consensus, resembling an artificial echo chamber. From a sociological perspective, this tendency suggests that large-scale multi-agent dialogues by LLMs may exaggerate conformity and under-represent minority perspectives, making them less realistic and potentially misleading as tools for studying collective behavior in large groups.

\subsection{Preference data analysis}

\begin{figure*}[!t]
  \centering
    \includegraphics[width=\columnwidth]{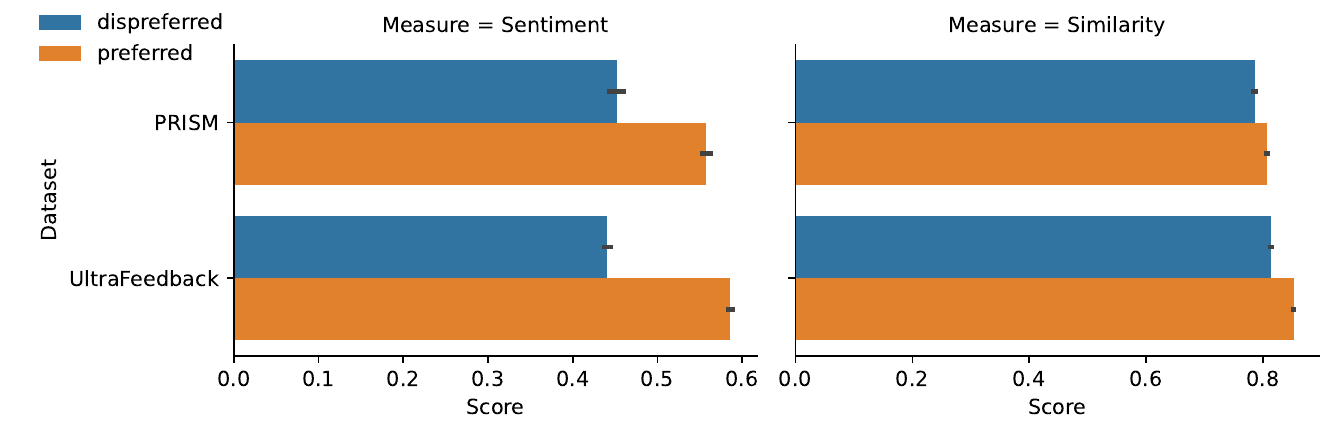}
    \caption{\textbf{Sentiment and semantic similarity in preference datasets.} 
    Average sentiment scores (\textit{left}) and semantic similarity (\textit{right}) for preferred and dispreferred responses in PRISM and UltraFeedback. Preferred outputs exhibit both higher positivity and stronger semantic alignment. All error bars represent 95\% confidence. The PRISM dataset contains 39,634 dispreferred and 97,108 preferred responses (total $N=136,742$). The UltraFeedback dataset contains 124,504 dispreferred and 387,224 preferred responses (total $N=511,728$).}
  \label{figure-f6}
\end{figure*}

To better understand the potential origins of the biases observed in simulated group interactions, we examined sentiment score and semantic similarity distributions in two widely used preference datasets: PRISM and UltraFeedback (Fig.~\ref{figure-f6}). Across both datasets, we observed a consistent skew in sentiment: preferred responses showed significantly higher average positivity than dispreferred ones ($p<0.001$), with PRISM exhibiting a larger gap than UltraFeedback. This suggests that reinforcement learning from human feedback (RLHF) pipelines may implicitly favor emotionally positive utterances, which could result in the over-positive tone of LLM-generated conversations.

Similarly, semantic similarity analyses revealed that preferred responses were more semantically aligned with their prompts compared to dispreferred ones ($p<0.001$). PRISM again displayed a larger gap. This pattern indicates that models trained on these datasets may be implicitly encouraged to produce more on-topic and semantically redundant responses, contributing to the higher similarity we observed in simulated chats. 

Together, these findings suggest that the affective and structural biases present in preference data may play a direct role in shaping the behavioral patterns of LLMs in social contexts.

\section{Discussion}

In this study, we introduced a systematic framework to examine how LLMs simulate human social interactions and collective behavior. 
We found that, rather than mirroring the complexity and variability of authentic social discourse, LLM-based group conversations tend to produce idealized, emotionally positive, and conflict-free societies. This pattern, which we term ``Utopian illusion'', reflects a systematic bias toward coherence, harmony, and prosociality in model-generated conversations. These findings highlight a fundamental divergence between optimization for socially acceptable language and the realities of human social behavior, suggesting that current LLMs simulate how people wish to behave rather than how they actually do.

This tendency toward idealized and harmonious interaction is analogous to the well-documented phenomenon of social desirability bias (SDB) in human psychology, in which individuals present themselves in socially acceptable and emotionally positive ways \cite{paulhus1984two, paulhus1991measurement, gower2022cognitive}. While LLMs do not possess intentions or self-presentation motives, their language outputs reflect the socially normative, cooperative, and emotionally stable patterns embedded in human texts. In multi-agent settings, this bias is further amplified: each agent responds to already desirable utterances from others, reinforcing conformity and politeness through recursive feedback.  As the group grows, this loop produces increasingly uniform and polished conversations, highlighting a key difference from authentic human interactions.

Within this SDB framework, we identify three specific social cognitive biases that consistently emerge in LLM-simulated group dialogues: social role bias, primacy effect, and positivity bias.

Firstly, the \textbf{social role bias} reflects how LLMs reproduce the prestige hierarchies embedded in their training data. LLMs consistently overproduce high-status, idealized social roles such as professionals, managers, and technical workers, while underrepresenting the majority of the global workforce, particularly those in elementary and agricultural sectors. These generated roles align with aspirational identities people hope to achieve, rather than reflecting real occupational distributions. Analysis of the Google Books Ngram Dataset suggests that this bias stems from the reporting bias of social roles in training corpora, where higher-status roles are more prominently represented in the text corpora used to train LLMs.

Secondly, we observe the \textbf{primacy effect}, a cognitive phenomenon in which information introduced early in a conversation shapes the dialogue \cite{murdock1962serial}. This effect is evident in both semantic similarity and keyword persistence analyses. LLM-generated conversations exhibit significantly higher inter- and intra-agent semantic similarity than human interactions, indicating a stronger tendency to stay on the same topics. The keyword persistence analysis further shows that keywords introduced early, particularly in the first utterance, carry more weight and exert more influence throughout the conversation in LLM-generated conversations. Once a topic (such as ``park'' in Fig. \ref{figure-f0}) is introduced in the early turns, it often becomes an anchor that shapes the rest of the discussion. This creates what we call a ``hollow spiral'', where the conversation repeatedly circles around the same themes without meaningful development. As a result, group viewpoints converge over time, giving the illusion of consensus without genuine diversity. In previous studies, similar tendencies have been observed in single-turn instruction-following tasks with ChatGPT and other LLMs, where models prefer options presented earlier in the prompt \cite{wang2023primacy, jumelet2024language, hamalainen2025psychology}. Our results extend this observation to multi-turn, multi-agent settings. 

Finally, the \textbf{positivity bias} provides the clearest evidence of social desirability. LLMs generate utterances with significantly higher sentiment scores than humans, expressing more agreement, praise, and optimism. Linguistic markers of disagreements and negations are noticeably reduced, making conversations polite but emotionally flat. This effect becomes even stronger in larger chat groups, where average sentiment increases sharply as the number of participants grows. We believe this pattern arises from LLMs' preference for safe, socially acceptable outputs. While such behavior may be desirable in human-AI interactions, it undermines the authenticity of simulated group dynamics. 

These tendencies align with prior observations that LLMs can exhibit sycophantic behavior, i.e., excessive agreement and flattery in both propositional and social contexts \cite{fanous2025syceval, malmqvist2024sycophancy, sharma2023towards, cheng2025social}. In propositional tasks, models often mirror users’ stated beliefs, even when these beliefs are inaccurate. In social interactions, they tend to preserve users’ face through emotional validation or moral endorsement. These behaviors may be interpreted as a form of social desirability bias: LLMs are systematically optimized to produce outputs that humans find agreeable, reinforcing tendencies toward conformity and positive emotions.

These biases originate from the very processes that make LLMs socially acceptable: the alignment-driven optimization. The idealized behavior is not only a mirror of how people tend to present themselves in public discourse, but also of how people prefer to be treated, especially by artificial agents. Through reinforcement learning from human feedback (RLHF) and preference tuning \cite{ouyang2022training}, models are rewarded for producing responses that humans find agreeable, polite, and emotionally stable, optimized not for truthfulness or diversity, but for likability and comfort. Over time, this objective implicitly leads to a form of social desirability bias. Our analysis of preference datasets supports this view: preferred responses consistently display higher sentiment scores and greater semantic coherence than dispreferred ones. Human annotators tend to favor outputs that affirm rather than challenge, reflecting a collective preference for social smoothness over cognitive dissonance. In this way, the alignment process systematically narrows the expressive range of models, aligning them with the idealized self-presentation patterns that characterize SDB.

Interestingly, models optimized for reasoning, such as GPT-O3 and DeepSeek-Reasoner, exhibit weaker sentiment inflation and greater thematic diversity. This suggests that when models are encouraged to engage in deeper reasoning rather than merely aiming to ``please'', they can exhibit more authentic behavior. This supports the view that reasoning-optimized models are more rational \cite{wang2025adaptive}. Moreover, this observation points to a potential strategy for mitigating SDB: introducing structured reasoning steps may disrupt the automatic reinforcement of socially desirable patterns, allowing models to better handle disagreement and ambiguity.

These findings raise a broader question: what kind of social alignment do we want from AI systems? If models are optimized primarily for socially desirable and conflict-averse outputs, they may fail to capture the richness, variability, and complexity inherent in human social life. In social simulations and social science experiments, such biases could generate illusions of consensus that mislead our understanding of human behavior, masking the roles of disagreement, inequality, and emotional diversity. For example, if we try to predict phenomena such as panic buying during a public health crisis or the spread of misinformation online, a ``utopian'' model may suggest that everyone behaves rationally, calmly discussing issues without panic or conflict. In reality, such models would underestimate the complexity and risks in actual social contexts, and relying on them for policy decisions could lead to seriously misguided outcomes. This poses significant risks to the accuracy of scientific insights and to the safe deployment of AI in real-world settings \cite{tsvetkova2024new}. 

To move beyond Utopian alignment, we must design AI systems that tolerate discomfort and disagreement, engage with moral and emotional complexity, and represent the heterogeneity that characterizes authentic human interaction \cite{bail2024can}. Approaches such as antagonistic AI, which encourage constructive disagreement and critical engagement, may help achieve these objectives \cite{cai2024antagonistic}. The goal is not to build agents that make us comfortable, but ones that help us see ourselves more clearly, including the aspects we prefer to avoid. Realizing this requires new forms of training data, generation objectives, and evaluation criteria that move beyond measures of likability and politeness toward metrics capturing authenticity, diversity, and alignment with the full spectrum of human social behavior.

Building on these principles, several practical strategies can help mitigate social desirability bias in LLMs and promote more authentic and diverse social simulations. One approach involves reasoning-oriented models that engage in structured and deliberate inference, such as chain-of-thought reasoning, which in our experiments has been associated with reduced sentiment inflation and higher conversational diversity. Incorporating diverse and emotionally unfiltered texts into training corpora allows models to capture a wider spectrum of human expression. Generation objectives can be adapted to reward disagreement and divergence, and fine-tuning on datasets reflecting real-world social variability may further enhance authenticity. Finally, evaluation metrics should move beyond social acceptability or politeness, incorporating measures like social authenticity and diversity.

While this study provides a valuable step toward understanding how LLMs simulate social interaction and collective behavior, several limitations should be noted. First, the analysis focused on eight representative models from three major families (GPT, Llama, DeepSeek), which may not capture the full diversity of architectures and training paradigms. Further experiments should extend to additional model families and architectures. Second, the group chat setting, though useful for isolating core interaction elements, represents a simplified and structured environment. Non-verbal cues, shared histories, physical contexts, spontaneous topic shifts, and subgroup dynamics were not represented, and the cyclic turn-taking design may constrain natural phenomena such as interruptions or selective participation. Moreover, the interaction setting was tension-free and non-goal-oriented, leaving open the question of how biases might emerge under resource constraints, competition, or collective decision-making. Third, evaluation relied on established quantitative metrics (e.g., cosine similarity, VADER sentiment, LIWC, KeyBERT), which, while systematic, cannot fully reflect the complexity and subtle aspects of social behaviors. Finally, although the analogy between human social desirability bias and LLM alignment bias offers a useful conceptual framework, the underlying mechanisms remain correlational rather than causal. Future work should disentangle how specific training or alignment procedures give rise to SDB-like tendencies. Despite these limitations, the results provide a foundation for future research to evaluate and enhance the complexity and variability of social interactions in LLM-based simulations.

\section{Methods}

\subsection{Models}
We use 3 widely recognized families of LLMs in our study: OpenAI's GPT, Meta's Llama, and DeepSeek. 

GPT models are state-of-the-art LLMs developed by OpenAI, widely recognized for their fluency, contextual coherence, and nuanced language generation. Starting with ChatGPT \cite{chatgpt}, these models have demonstrated strong abilities in producing fluent, human-like conversations. In our study, we use three versions of GPT model: GPT-4o (\textit{gpt-4o}) \cite{hurst2024gpt}, a conversational model optimized for alignment with human preferences; GPT-o3 (\textit{o3}) \cite{o3}, a reasoning-focused model; and InstructGPT (\textit{gpt-3.5-turbo-instruct}) \cite{ouyang2022training}, an earlier version trained for instruction following but with limited alignment to human values.

Llama models \cite{touvron2023llama} are open-source LLMs developed by Meta AI. Available in various sizes, they are optimized for open-domain conversations and demonstrate strong performance in both general reasoning and social language use. In our study, we used Llama 3.1 (8B) \cite{llama3.1}, Llama 3.3 (70B), and Llama 4 (17B-16E) \cite{llama4}. 

DeepSeek models \cite{bi2024deepseek} are competitive LLMs developed in China, trained on a large-scale corpus covering both Chinese and English. Despite being trained with relatively lower computational costs, DeepSeek achieves high-quality language generation performance. In our experiments, we used two variants: DeepSeek-Chat \cite{liu2024deepseek}, optimized for conversational interaction, and DeepSeek-Reasoner \cite{guo2025deepseek}, designed for enhanced reasoning abilities.

In our experiments, we accessed the GPT and DeepSeek models through their APIs, while the Llama models were run locally on two RTX 4090 GPUs. All generations were performed using the models’ default temperature and sampling configurations to ensure consistency across experiments.

\subsection{Human conversation datasets}
For comparison with simulated conversations, we use two human-human dialogue datasets: TopicalChat and PersonaChat.

TopicalChat \cite{gopalakrishnan2023topical} is a human-human dialogue dataset designed to support knowledge-grounded conversations on diverse topics such as politics, science, entertainment, and sports. The dataset was collected via Amazon Mechanical Turk, where pairs of workers were provided with reading sets composed of curated Wikipedia passages, fun facts, and Washington Post articles about specific entities. These reading sets were either symmetric (shared knowledge) or asymmetric (varying degrees of information exposure), enabling natural conversational dynamics and implicit role shifts between teaching and learning. Conversations were required to be at least 20 turns long, and Turkers annotated each turn with sentiment, knowledge source, and message quality. This dataset provides a realistic reference for assessing the diversity, coherence, and topic adherence of multi-turn dialogues in both human and LLM-generated conversations. We use its training set in our experiments, which comprises 188,378 utterances across 8,628 dialogues.

PersonaChat \cite{zhang2018personalizing} is a crowd-sourced human-human dialogue dataset aimed at enabling more engaging and personalized chit-chat by grounding conversations in character-driven personas. The data collection process involved Amazon Mechanical Turk workers creating persona profiles made up of five natural-sounding sentences. To reduce the impact of lexical overlap and increase task difficulty, each persona was later rewritten by different workers into semantically related but lexically distinct versions. In the dialogue phase, pairs of workers were randomly assigned different personas and instructed to engage in a natural conversation while subtly incorporating their character's traits. This structure encourages mutual exploration of personal attributes and social interaction. The final dataset comprises 162,064 utterances across 10,907 dialogues, and we use its training set in our experiments.

While TopicalChat and PersonaChat reflect text-based, task-oriented interactions among crowd workers and thus do not capture more complex, real-world conversational settings, these datasets still provide a robust benchmark for assessing dialogue coherence, topical adherence, and multi-turn dynamics. Importantly, the clear gap observed between LLM-generated dialogues and these human baselines underscores that our main findings are robust, even considering the specific nature of these datasets.

\subsection{Preference datasets}

To investigate potential sources of the cognitive biases observed in LLM-generated group interactions, we analyzed two preference datasets commonly used in post-training and alignment: PRISM \cite{kirk2024prism} and UltraFeedback \cite{cui2024ultrafeedback}. Both datasets provide scalar feedback scores rather than binary labels. PRISM scores range from 0–100, and UltraFeedback scores range from 0–10. For our analysis, we used the midpoint of each scale (50 for PRISM and 5 for UltraFeedback) as a threshold: responses with scores above the midpoint were treated as \textit{preferred}, and those below as \textit{dispreferred}. By comparing sentiment and semantic similarity patterns in these preferred and dispreferred responses with those observed in our simulated LLM group conversations, we assess the extent to which post-training preference signals may contribute to the amplification of positivity bias and semantic redundancy in multi-agent dialogue.

\subsection{Experimental design}

Our experimental procedure consisted of three stages: selecting a chat topic, assigning social roles to each LLM agent, and conducting the conversation.

Chat topics were drawn from three sources: (i) generated by the LLM itself (e.g., \textit{``What if all technology stopped working for a week - how would people adapt?''}, the prompt for generating topics is shown in Appendix \ref{secA1}); (ii) constructed from three keywords randomly sampled from the TopicalChat dataset (e.g., \textit{``Telephone, Google, Human''}); or (iii) selected from debate topics used in \cite{salvi2025conversational} (e.g., \textit{``Is government surveillance necessary for national security?''}).
For (i) and (ii), we sample 10 topics for each experimental setting, and for (iii), we use all 30 debate topics. So there were 10+10+30=50 topics for each experimental setting.

Social roles were assigned based on two sources: (i) roles generated by the LLM itself (e.g., \textit{``Person 1 is a 34-year-old woman living in a suburban area of a mid-sized city. She identifies as a middle-class working professional and is employed as a marketing coordinator at a local company \dots''}, the prompt for generating social roles is shown in Appendix \ref{secA2}); or (ii) randomly sampled from the PersonaChat dataset (e.g., \textit{``People say I have a cute laugh; I am still in love with my ex-boyfriend; I love to cook for my family and friends; I work in a publishing building; I am a female and love to be surrounded by males.''})

We varied the number of participants in the simulated chat from 2, 3, 4, 5, and 8. For experiments with the Llama 3.1 model, we further extended the number of participants to 12, 16, 24, and 32.

After the chat topic and social roles were defined, the conversation commenced. The conversation was initiated by Person 1 and followed a cyclic turn-taking scheme, looping through each agent in order (e.g., Person 1 $\rightarrow$ Person 2 $\rightarrow$ Person 3 $\rightarrow$ Person 1 $\rightarrow$ \dots).
Each agent was prompted to adhere to its assigned social role and to contribute a brief response of two to three sentences in the group chat, considering the chat topic and the conversation history thus far (see Appendix \ref{secA3} and \ref{secA4} for the prompts). At each turn, agents could choose either to continue the discussion or remain silent if they had nothing to contribute. Agents who chose to remain silent did not have their turn recorded in the chat history. The conversation terminated once a predefined maximum number of utterances was reached. Specifically, we set the maximum number of utterances to 30 for groups with up to 5 participants, 50 for groups of 8, 70 for groups of 12, 90 for groups of 16, 120 for groups of 24, and 150 for groups of 32.

In total, we systematically varied four independent variables: (i) \textbf{Model}, 8 LLMs covering unaligned, instruction-tuned, and reasoning-optimized variants; (ii) \textbf{Source of conversation topics}, from TopicalChat, debate topics~\cite{salvi2025conversational}; (iii) \textbf{Source of social roles}, either predefined or LLM-generated; (iv) \textbf{Number of participants} in the conversation, or LLM-generated. We conducted 8 (models) × 50 (topics) × 2 (social role sources) × 5 (number of participants) = 4000 conversations. For Llama 3.1, there are another 50 (topics) × 2 (social role sources) × 4 (number of participants) = 400 conversations. So there are 4400 conversations in total in our experiments.

We designed this experimental paradigm because it allows for a high degree of control over conversational dynamics, enhancing internal validity. By conducting sequential, text-based chats, we ensure that any observed patterns or biases can be attributed to the LLMs themselves and the structural properties of the conversation, rather than external factors such as tone of voice, gestures, or prior knowledge. This design provides a precise framework for studying how LLMs generate dialogue while adhering to assigned social roles. It is not intended to generalize to all of social simulation, but specifically to discourse-dependent social dynamics such as committee meetings, deliberation forums, or online groups, where a history of shared text is the primary input for decision-making.

\subsection{Evaluation Metrics}

To measure the differences between LLM-simulated and human conversations, we evaluated them along five dimensions:

(i) \textbf{Role distribution}, defined as the distribution of social roles generated by the LLMs compared to real-world occupational distributions. We classified each model-generated social role into one of the ten major occupational groups (coded 0–9) defined by the International Standard Classification of Occupations (ISCO) from the International Labour Organization (ILO). To perform the classification, we used the GLM-Z1-flash model, a reasoning-oriented LLM developed by Zhipu AI with a free API, which assigned an ISCO occupation code based on the role descriptions and example occupations from each major group (see prompt in Appendix \ref{secA5}). This model was chosen because it achieves a good balance between cost and performance. We ran the classification three times and used the majority vote as the final result. To ensure the reliability of the automatic classification method, we also randomly sampled 100 social roles, manually annotated their occupational groups, and compared these labels with the automatic classification results. The agreement between the manual and automatic classifications was 83\%. Finally, we compared the resulting occupational distribution to the real-world global occupational structure in 2023, as reported in the ILO's World Employment and Social Outlook. 

(ii) \textbf{Conversational similarity}, quantified along two dimensions: \textit{intra-agent similarity}, measuring the semantic similarity between an agent's current utterance and its own previous utterance; and \textit{inter-agent similarity}, measuring the semantic similarity between the current utterance and the immediately preceding utterance in the conversation, regardless of speaker. Both similarity measures were computed using embeddings from the BERT-large model \cite{devlin-etal-2019-bert}, and quantified as the cosine similarity between sentence representations. 

(iii) \textbf{Keyword Persistence}. To quantify keyword persistence and examine the influence of early conversational content, we computed a normalized keyword weight for each utterance based on the output of the KeyBERT model \cite{grootendorst2020keybert}. For every conversation, we extracted keywords and their associated importance weights for each utterance. We then aggregated keyword weights across the full conversation to compute a global importance score for each unique keyword. Specifically, for a given keyword, we summed its weights across all utterances and divided by the total number of utterances in the conversation, yielding a normalized weight that reflected both its prominence and persistence throughout the dialogue.
To track the effect of early keyword introduction, we recorded the first turn in which each keyword appeared. For every keyword-utterance pair where the keyword appeared for the first time, we assigned the normalized weight to that turn. Keywords that reappeared in later turns were not double-counted. This allowed us to associate a persistent weight with the utterance where the keyword originated, capturing how long and strongly the keyword influenced the subsequent conversation.

(iv) \textbf{Emotional tone}, assessed via sentiment analysis of each utterance using the Valence Aware Dictionary and sEntiment Reasoner (VADER) \cite{ahuja2017clustering}.

(v) \textbf{Linguistic patterns}, analyzed using the Linguistic Inquiry and Word Count (LIWC) tool \cite{tausczik2010psychological, pennebaker2015development}. LIWC quantifies the prevalence of psychologically and socially meaningful language categories, including function words (e.g., pronouns, articles), affective processes (e.g., positive or negative emotion), cognitive processes (e.g., insight, causation), social processes (e.g., family, friends), and other dimensions relevant to communication style. For each utterance, we computed the relative frequency of words in LIWC categories, allowing us to characterize the linguistic and psychological tendencies of model-generated conversations and compare them with human communication patterns.

\bibliographystyle{bst/sn-aps}
\bibliography{sn-bibliography}

\begin{thebibliography}{10}
\providecommand{\url}[1]{{#1}}
\providecommand{\urlprefix}{URL }
\providecommand{\doi}[1]{\url{https://doi.org/#1}}


\bibitem{jimenez2021experimental}
M.~Jim{\'e}nez-Buedo, F.~Russo, Experimental practices and objectivity in the social sciences: re-embedding construct validity in the internal--external validity distinction.
\newblock Synthese \textbf{199}(3), 9549--9579 (2021)

\bibitem{gilbert2000build}
N.~Gilbert, P.~Terna, How to build and use agent-based models in social science.
\newblock Mind \& Society \textbf{1}(1), 57--72 (2000)

\bibitem{davidsson2002agent}
P.~Davidsson, Agent based social simulation: A computer science view.
\newblock Journal of artificial societies and social simulation \textbf{5}(1) (2002)

\bibitem{fan2024agent}
R.~Fan, Q.~Yao, R.~Chen, R.~Qian, Agent-based simulation model of panic buying behavior in urban public crisis events: A social network perspective.
\newblock Sustainable Cities and Society \textbf{100}, 105002 (2024)

\bibitem{fu2021propagation}
P.~Fu, B.~Jing, T.~Chen, C.~Xu, J.~Yang, G.~Cong, Propagation model of panic buying under the sudden epidemic.
\newblock Frontiers in public health \textbf{9}, 675687 (2021)

\bibitem{pan2007multi}
X.~Pan, C.S. Han, K.~Dauber, K.H. Law, A multi-agent based framework for the simulation of human and social behaviors during emergency evacuations.
\newblock Ai \& Society \textbf{22}(2), 113--132 (2007)

\bibitem{butler2024mis}
L.H. Butler, P.~Lamont, D.L.Y. Wan, T.~Prike, M.~Nasim, B.~Walker, N.~Fay, U.K. Ecker, The (mis) information game: A social media simulator.
\newblock Behavior Research Methods \textbf{56}(3), 2376--2397 (2024)

\bibitem{tambuscio2015fact}
M.~Tambuscio, G.~Ruffo, A.~Flammini, F.~Menczer, \emph{Fact-checking effect on viral hoaxes: A model of misinformation spread in social networks}, in \emph{Proceedings of the 24th international conference on World Wide Web} (2015), pp. 977--982

\bibitem{hofman2021integrating}
J.M. Hofman, D.J. Watts, S.~Athey, F.~Garip, T.L. Griffiths, J.~Kleinberg, H.~Margetts, S.~Mullainathan, M.J. Salganik, S.~Vazire, et~al., Integrating explanation and prediction in computational social science.
\newblock Nature \textbf{595}(7866), 181--188 (2021)

\bibitem{conte2014agent}
R.~Conte, M.~Paolucci, On agent-based modeling and computational social science.
\newblock Frontiers in psychology \textbf{5}, 668 (2014)

\bibitem{chandramouli2024workflow}
S.~Chandramouli, D.~Shi, A.~Putkonen, S.~De~Peuter, S.~Zhang, J.~Jokinen, A.~Howes, A.~Oulasvirta, A workflow for building computationally rational models of human behavior.
\newblock Computational Brain \& Behavior \textbf{7}(3), 399--419 (2024)

\bibitem{chatgpt}
OpenAI.
\newblock Introducing chatgpt (2022).
\newblock \urlprefix\url{https://openai.com/blog/chatgpt}.
\newblock 2022, Nov 30

\bibitem{touvron2023llama}
H.~Touvron, T.~Lavril, G.~Izacard, X.~Martinet, M.A. Lachaux, T.~Lacroix, B.~Rozière, N.~Goyal, E.~Hambro, F.~Azhar, A.~Rodriguez, A.~Joulin, E.~Grave, G.~Lample, Llama: Open and efficient foundation language models.
\newblock arXiv preprint arXiv:2302.13971  (2023)

\bibitem{guo2025deepseek}
D.~Guo, D.~Yang, H.~Zhang, J.~Song, R.~Zhang, R.~Xu, Q.~Zhu, S.~Ma, P.~Wang, X.~Bi, et~al., Deepseek-r1: Incentivizing reasoning capability in llms via reinforcement learning.
\newblock arXiv preprint arXiv:2501.12948  (2025)

\bibitem{strachan2024testing}
J.W. Strachan, D.~Albergo, G.~Borghini, O.~Pansardi, E.~Scaliti, S.~Gupta, K.~Saxena, A.~Rufo, S.~Panzeri, G.~Manzi, et~al., Testing theory of mind in large language models and humans.
\newblock Nature Human Behaviour \textbf{8}(7), 1285--1295 (2024)

\bibitem{li2023camel}
G.~Li, H.~Hammoud, H.~Itani, D.~Khizbullin, B.~Ghanem, Camel: Communicative agents for" mind" exploration of large language model society.
\newblock Advances in Neural Information Processing Systems \textbf{36}, 51991--52008 (2023)

\bibitem{kosinski2024evaluating}
M.~Kosinski, Evaluating large language models in theory of mind tasks.
\newblock Proceedings of the National Academy of Sciences \textbf{121}(45), e2405460121 (2024)

\bibitem{piao2025agentsociety}
J.~Piao, Y.~Yan, J.~Zhang, N.~Li, J.~Yan, X.~Lan, Z.~Lu, Z.~Zheng, J.Y. Wang, D.~Zhou, et~al., Agentsociety: Large-scale simulation of llm-driven generative agents advances understanding of human behaviors and society.
\newblock arXiv preprint arXiv:2502.08691  (2025)

\bibitem{dillion2023can}
D.~Dillion, N.~Tandon, Y.~Gu, K.~Gray, Can ai language models replace human participants?
\newblock Trends in Cognitive Sciences \textbf{27}(7), 597--600 (2023).
\newblock \doi{https://doi.org/10.1016/j.tics.2023.04.008}

\bibitem{griffin2023susceptibility}
L.D. Griffin, B.~Kleinberg, M.~Mozes, K.T. Mai, M.~Vau, M.~Caldwell, A.~Marvor-Parker, Susceptibility to influence of large language models.
\newblock arXiv preprint arXiv:2303.06074  (2023)

\bibitem{argyle2023out}
L.P. Argyle, E.C. Busby, N.~Fulda, J.R. Gubler, C.~Rytting, D.~Wingate, Out of one, many: Using language models to simulate human samples.
\newblock Political Analysis \textbf{31}(3), 337--351 (2023)

\bibitem{aher2023using}
G.V. Aher, R.I. Arriaga, A.T. Kalai, \emph{Using large language models to simulate multiple humans and replicate human subject studies}, in \emph{International conference on machine learning} (PMLR, 2023), pp. 337--371

\bibitem{park2022social}
J.S. Park, L.~Popowski, C.~Cai, M.R. Morris, P.~Liang, M.S. Bernstein, \emph{Social simulacra: Creating populated prototypes for social computing systems}, in \emph{Proceedings of the 35th Annual ACM Symposium on User Interface Software and Technology} (2022), pp. 1--18

\bibitem{hagendorff2023human}
T.~Hagendorff, S.~Fabi, M.~Kosinski, Human-like intuitive behavior and reasoning biases emerged in large language models but disappeared in chatgpt.
\newblock Nature Computational Science \textbf{3}(10), 833--838 (2023)

\bibitem{lehr2025kernels}
S.A. Lehr, K.S. Saichandran, E.~Harmon-Jones, N.~Vitali, M.R. Banaji, Kernels of selfhood: Gpt-4o shows humanlike patterns of cognitive dissonance moderated by free choice.
\newblock Proceedings of the National Academy of Sciences \textbf{122}(20), e2501823122 (2025)

\bibitem{zhou2023sotopia}
X.~Zhou, H.~Zhu, L.~Mathur, R.~Zhang, H.~Yu, Z.~Qi, L.P. Morency, Y.~Bisk, D.~Fried, G.~Neubig, et~al., Sotopia: Interactive evaluation for social intelligence in language agents.
\newblock arXiv preprint arXiv:2310.11667  (2023)

\bibitem{wang2024sotopia}
R.~Wang, H.~Yu, W.~Zhang, Z.~Qi, M.~Sap, Y.~Bisk, G.~Neubig, H.~Zhu, \emph{SOTOPIA-$\pi$: Interactive Learning of Socially Intelligent Language Agents}, in \emph{Proceedings of the 62nd Annual Meeting of the Association for Computational Linguistics (Volume 1: Long Papers)} (2024), pp. 12912--12940

\bibitem{tang2024gensim}
J.~Tang, H.~Gao, X.~Pan, L.~Wang, H.~Tan, D.~Gao, Y.~Chen, X.~Chen, Y.~Lin, Y.~Li, et~al., Gensim: A general social simulation platform with large language model based agents.
\newblock arXiv preprint arXiv:2410.04360  (2024)

\bibitem{mou2024individual}
X.~Mou, X.~Ding, Q.~He, L.~Wang, J.~Liang, X.~Zhang, L.~Sun, J.~Lin, J.~Zhou, X.~Huang, et~al., From individual to society: A survey on social simulation driven by large language model-based agents.
\newblock arXiv preprint arXiv:2412.03563  (2024)

\bibitem{gao2024large}
C.~Gao, X.~Lan, N.~Li, Y.~Yuan, J.~Ding, Z.~Zhou, F.~Xu, Y.~Li, Large language models empowered agent-based modeling and simulation: A survey and perspectives.
\newblock Humanities and Social Sciences Communications \textbf{11}(1), 1--24 (2024)

\bibitem{yang2024oasis}
Z.~Yang, Z.~Zhang, Z.~Zheng, Y.~Jiang, Z.~Gan, Z.~Wang, Z.~Ling, J.~Chen, M.~Ma, B.~Dong, et~al., Oasis: Open agents social interaction simulations on one million agents.
\newblock arXiv preprint arXiv:2411.11581  (2024)

\bibitem{park2023generative}
J.S. Park, J.~O'Brien, C.J. Cai, M.R. Morris, P.~Liang, M.S. Bernstein, \emph{Generative agents: Interactive simulacra of human behavior}, in \emph{Proceedings of the 36th annual acm symposium on user interface software and technology} (2023), pp. 1--22

\bibitem{gallegos2024bias}
I.O. Gallegos, R.A. Rossi, J.~Barrow, M.M. Tanjim, S.~Kim, F.~Dernoncourt, T.~Yu, R.~Zhang, N.K. Ahmed, Bias and fairness in large language models: A survey.
\newblock Computational Linguistics \textbf{50}(3), 1097--1179 (2024)

\bibitem{kotek2023gender}
H.~Kotek, R.~Dockum, D.~Sun, \emph{Gender bias and stereotypes in large language models}, in \emph{Proceedings of the ACM collective intelligence conference} (2023), pp. 12--24

\bibitem{navigli2023biases}
R.~Navigli, S.~Conia, B.~Ross, Biases in large language models: origins, inventory, and discussion.
\newblock ACM Journal of Data and Information Quality \textbf{15}(2), 1--21 (2023)

\bibitem{plaza2024angry}
F.~Plaza-del Arco, A.~Curry, A.C. Curry, G.~Abercrombie, D.~Hovy, \emph{Angry Men, Sad Women: Large Language Models Reflect Gendered Stereotypes in Emotion Attribution}, in \emph{Proceedings of the 62nd Annual Meeting of the Association for Computational Linguistics (Volume 1: Long Papers)} (2024), pp. 7682--7696

\bibitem{demidova2024john}
A.~Demidova, H.~Atwany, N.~Rabih, S.~Sha’ban, M.~Abdul-Mageed, \emph{John vs. ahmed: Debate-induced bias in multilingual LLMs}, in \emph{Proceedings of The Second Arabic Natural Language Processing Conference} (2024), pp. 193--209

\bibitem{wan2024white}
Y.~Wan, K.W. Chang, White men lead, black women help? benchmarking language agency social biases in llms.
\newblock arXiv preprint arXiv:2404.10508  (2024)

\bibitem{plaza2024divine}
F.~Plaza-del Arco, A.~Curry, S.~Paoli, A.C. Curry, D.~Hovy, \emph{Divine LLaMAs: Bias, Stereotypes, Stigmatization, and Emotion Representation of Religion in Large Language Models}, in \emph{Findings of the Association for Computational Linguistics: EMNLP 2024} (2024), pp. 4346--4366

\bibitem{bian2024influence}
N.~Bian, H.~Lin, P.~Liu, Y.~Lu, C.~Zhang, B.~He, X.~Han, L.~Sun, Influence of external information on large language models mirrors social cognitive patterns.
\newblock IEEE Transactions on Computational Social Systems  (2024)

\bibitem{koo2024benchmarking}
R.~Koo, M.~Lee, V.~Raheja, J.I. Park, Z.M. Kim, D.~Kang, \emph{Benchmarking Cognitive Biases in Large Language Models as Evaluators}, in \emph{Findings of the Association for Computational Linguistics ACL 2024} (2024), pp. 517--545

\bibitem{itzhak2024instructed}
I.~Itzhak, G.~Stanovsky, N.~Rosenfeld, Y.~Belinkov, Instructed to bias: Instruction-tuned language models exhibit emergent cognitive bias.
\newblock Transactions of the Association for Computational Linguistics \textbf{12}, 771--785 (2024)

\bibitem{chen2025manager}
Y.~Chen, S.N. Kirshner, A.~Ovchinnikov, M.~Andiappan, T.~Jenkin, A manager and an ai walk into a bar: does chatgpt make biased decisions like we do?
\newblock Manufacturing \& Service Operations Management  (2025)

\bibitem{hu2025generative}
T.~Hu, Y.~Kyrychenko, S.~Rathje, N.~Collier, S.~van~der Linden, J.~Roozenbeek, Generative language models exhibit social identity biases.
\newblock Nature Computational Science \textbf{5}(1), 65--75 (2025)

\bibitem{hewitt2024predicting}
L.~Hewitt, A.~Ashokkumar, I.~Ghezae, R.~Willer.
\newblock Predicting results of social science experiments using large language models (2024).
\newblock \urlprefix\url{https://samim.io/dl/Predicting\%20results\%20of\%20social\%20science\%20experiments\%20using\%20large\%20language\%20models.pdf}.
\newblock Stanford University and New York University, August 8, 2024

\bibitem{cfeae38342824620aafa35192bd0fda8}
M.~Sharma, M.~Tong, T.~Korbak, D.~Duvenaud, A.~Askell, S.R. Bowman, N.~Cheng, E.~Durmus, Z.~Hatfield-Dodds, S.R. Johnston, et~al., Towards understanding sycophancy in language models.
\newblock arXiv preprint arXiv:2310.13548  (2023)

\bibitem{ouyang2022training}
L.~Ouyang, J.~Wu, X.~Jiang, D.~Almeida, C.~Wainwright, P.~Mishkin, C.~Zhang, S.~Agarwal, K.~Slama, A.~Ray, et~al., Training language models to follow instructions with human feedback.
\newblock Advances in neural information processing systems \textbf{35}, 27730--27744 (2022)

\bibitem{askell2021general}
A.~Askell, Y.~Bai, A.~Chen, D.~Drain, D.~Ganguli, T.~Henighan, A.~Jones, N.~Joseph, B.~Mann, N.~DasSarma, et~al., A general language assistant as a laboratory for alignment.
\newblock arXiv preprint arXiv:2112.00861  (2021)

\bibitem{kamath2024tuning}
U.~Kamath, K.~Keenan, G.~Somers, S.~Sorenson, in \emph{Large Language Models: A Deep Dive: Bridging Theory and Practice} (Springer, 2024), pp. 177--218

\bibitem{fontana2025nicer}
N.~Fontana, F.~Pierri, L.M. Aiello, \emph{Nicer Than Humans: How Do Large Language Models Behave in the Prisoner's Dilemma?}, in \emph{Proceedings of the International AAAI Conference on Web and Social Media}, vol.~19 (2025), pp. 522--535

\bibitem{ashery2025emergent}
A.F. Ashery, L.M. Aiello, A.~Baronchelli, Emergent social conventions and collective bias in llm populations.
\newblock Science Advances \textbf{11}(20), eadu9368 (2025)

\bibitem{demszky2023using}
D.~Demszky, D.~Yang, D.S. Yeager, C.J. Bryan, M.~Clapper, S.~Chandhok, J.C. Eichstaedt, C.~Hecht, J.~Jamieson, M.~Johnson, et~al., Using large language models in psychology.
\newblock Nature Reviews Psychology \textbf{2}(11), 688--701 (2023)

\bibitem{bail2024can}
C.A. Bail, Can generative ai improve social science?
\newblock Proceedings of the National Academy of Sciences \textbf{121}(21), e2314021121 (2024)

\bibitem{eagly2012social}
A.H. Eagly, W.~Wood, Social role theory.
\newblock Handbook of theories of social psychology \textbf{2}(9), 458--476 (2012)

\bibitem{chen2024persona}
J.~Chen, X.~Wang, R.~Xu, S.~Yuan, Y.~Zhang, W.~Shi, J.~Xie, S.~Li, R.~Yang, T.~Zhu, et~al., From persona to personalization: A survey on role-playing language agents.
\newblock arXiv preprint arXiv:2404.18231  (2024)

\bibitem{michel2011quantitative}
J.B. Michel, Y.K. Shen, A.P. Aiden, A.~Veres, M.K. Gray, G.B. Team, J.P. Pickett, D.~Hoiberg, D.~Clancy, P.~Norvig, et~al., Quantitative analysis of culture using millions of digitized books.
\newblock science \textbf{331}(6014), 176--182 (2011)

\bibitem{kirk2024prism}
H.R. Kirk, A.~Whitefield, P.~Rottger, A.M. Bean, K.~Margatina, R.~Mosquera-Gomez, J.~Ciro, M.~Bartolo, A.~Williams, H.~He, et~al., The prism alignment dataset: What participatory, representative and individualised human feedback reveals about the subjective and multicultural alignment of large language models.
\newblock Advances in Neural Information Processing Systems \textbf{37}, 105236--105344 (2024)

\bibitem{cui2024ultrafeedback}
G.~Cui, L.~Yuan, N.~Ding, G.~Yao, B.~He, W.~Zhu, Y.~Ni, G.~Xie, R.~Xie, Y.~Lin, et~al., \emph{ULTRAFEEDBACK: Boosting Language Models with Scaled AI Feedback}, in \emph{International Conference on Machine Learning} (PMLR, 2024), pp. 9722--9744

\bibitem{gopalakrishnan2023topical}
K.~Gopalakrishnan, B.~Hedayatnia, Q.~Chen, A.~Gottardi, S.~Kwatra, A.~Venkatesh, R.~Gabriel, D.~Hakkani-Tur, Topical-chat: Towards knowledge-grounded open-domain conversations.
\newblock arXiv preprint arXiv:2308.11995  (2023)

\bibitem{devlin-etal-2019-bert}
J.~Devlin, M.W. Chang, K.~Lee, K.~Toutanova, \emph{{BERT}: Pre-training of Deep Bidirectional Transformers for Language Understanding}, in \emph{Proceedings of the 2019 Conference of the North {A}merican Chapter of the Association for Computational Linguistics: Human Language Technologies, Volume 1 (Long and Short Papers)} (Association for Computational Linguistics, Minneapolis, Minnesota, 2019), pp. 4171--4186.
\newblock \doi{10.18653/v1/N19-1423}.
\newblock \urlprefix\url{https://aclanthology.org/N19-1423}

\bibitem{salvi2025conversational}
F.~Salvi, M.~Horta~Ribeiro, R.~Gallotti, R.~West, On the conversational persuasiveness of gpt-4.
\newblock Nature Human Behaviour  (2025)

\bibitem{grootendorst2020keybert}
M.~Grootendorst.
\newblock Keybert: Minimal keyword extraction with bert. (2020).
\newblock \doi{10.5281/zenodo.4461265}.
\newblock \urlprefix\url{https://doi.org/10.5281/zenodo.4461265}

\bibitem{murdock1962serial}
B.B. Murdock~Jr, The serial position effect of free recall.
\newblock Journal of experimental psychology \textbf{64}(5), 482 (1962)

\bibitem{ahuja2017clustering}
S.~Ahuja, G.~Dubey, \emph{Clustering and sentiment analysis on Twitter data}, in \emph{2017 2nd international conference on telecommunication and networks (TEL-NET)} (IEEE, 2017), pp. 1--5

\bibitem{tausczik2010psychological}
Y.R. Tausczik, J.W. Pennebaker, The psychological meaning of words: Liwc and computerized text analysis methods.
\newblock Journal of language and social psychology \textbf{29}(1), 24--54 (2010)

\bibitem{pennebaker2015development}
J.W. Pennebaker, R.L. Boyd, K.~Jordan, K.~Blackburn, The development and psychometric properties of liwc2015  (2015)

\bibitem{paulhus1984two}
D.L. Paulhus, Two-component models of socially desirable responding.
\newblock Journal of personality and social psychology \textbf{46}(3), 598 (1984)

\bibitem{paulhus1991measurement}
D.L. Paulhus, Measurement and control of response bias.  (1991)

\bibitem{gower2022cognitive}
T.~Gower, J.~Pham, E.N. Jouriles, D.~Rosenfield, H.J. Bowen, Cognitive biases in perceptions of posttraumatic growth: A systematic review and meta-analysis.
\newblock Clinical Psychology Review \textbf{94}, 102159 (2022)

\bibitem{wang2023primacy}
Y.~Wang, Y.~Cai, M.~Chen, Y.~Liang, B.~Hooi, \emph{Primacy Effect of ChatGPT}, in \emph{Proceedings of the 2023 Conference on Empirical Methods in Natural Language Processing} (2023), pp. 108--115

\bibitem{jumelet2024language}
J.~Jumelet, W.~Zuidema, A.~Sinclair, \emph{Do Language Models Exhibit Human-like Structural Priming Effects?}, in \emph{Findings of the Association for Computational Linguistics ACL 2024} (2024), pp. 14727--14742

\bibitem{hamalainen2025psychology}
M.~H{\"a}m{\"a}l{\"a}inen, \emph{On Psychology of AI--Does Primacy Effect Affect ChatGPT and Other LLMs?}, in \emph{Proceedings of the 5th International Conference on Natural Language Processing for Digital Humanities} (2025), pp. 209--213

\bibitem{fanous2025syceval}
A.~Fanous, J.~Goldberg, A.A. Agarwal, J.~Lin, A.~Zhou, R.~Daneshjou, S.~Koyejo, Syceval: Evaluating llm sycophancy.
\newblock arXiv preprint arXiv:2502.08177  (2025)

\bibitem{malmqvist2024sycophancy}
L.~Malmqvist, Sycophancy in large language models: Causes and mitigations.
\newblock arXiv preprint arXiv:2411.15287  (2024)

\bibitem{sharma2023towards}
M.~Sharma, M.~Tong, T.~Korbak, D.~Duvenaud, A.~Askell, S.R. Bowman, N.~Cheng, E.~Durmus, Z.~Hatfield-Dodds, S.R. Johnston, et~al., Towards understanding sycophancy in language models.
\newblock arXiv preprint arXiv:2310.13548  (2023)

\bibitem{cheng2025social}
M.~Cheng, S.~Yu, C.~Lee, P.~Khadpe, L.~Ibrahim, D.~Jurafsky, Social sycophancy: A broader understanding of llm sycophancy.
\newblock arXiv preprint arXiv:2505.13995  (2025)

\bibitem{wang2025adaptive}
M.~Wang, Y.~Li, H.~Wang, X.~Zhang, N.~Xu, B.~Wu, F.~Huang, H.~Yu, W.~Mao, Adaptive thinking via mode policy optimization for social language agents.
\newblock arXiv preprint arXiv:2505.02156  (2025)

\bibitem{tsvetkova2024new}
M.~Tsvetkova, T.~Yasseri, N.~Pescetelli, T.~Werner, A new sociology of humans and machines.
\newblock Nature Human Behaviour \textbf{8}(10), 1864--1876 (2024)

\bibitem{cai2024antagonistic}
A.~Cai, I.~Arawjo, E.L. Glassman, Antagonistic ai.
\newblock arXiv preprint arXiv:2402.07350  (2024)

\bibitem{hurst2024gpt}
A.~Hurst, A.~Lerer, A.P. Goucher, A.~Perelman, A.~Ramesh, A.~Clark, A.~Ostrow, A.~Welihinda, A.~Hayes, A.~Radford, et~al., Gpt-4o system card.
\newblock arXiv preprint arXiv:2410.21276  (2024)

\bibitem{o3}
OpenAI.
\newblock Introducing openai o3 and o4-mini (2025).
\newblock \urlprefix\url{https://openai.com/index/introducing-o3-and-o4-mini/}.
\newblock 2025, Apr 16

\bibitem{llama3.1}
Meta.
\newblock Introducing llama 3.1: Our most capable models to date (2024).
\newblock \urlprefix\url{https://ai.meta.com/blog/meta-llama-3-1/}.
\newblock 2024, Jul 23

\bibitem{llama4}
Meta.
\newblock The llama 4 herd: The beginning of a new era of natively multimodal ai innovation (2025).
\newblock \urlprefix\url{https://ai.meta.com/blog/llama-4-multimodal-intelligence/}.
\newblock 2025, Apr 5

\bibitem{bi2024deepseek}
X.~Bi, D.~Chen, G.~Chen, S.~Chen, D.~Dai, C.~Deng, H.~Ding, K.~Dong, Q.~Du, Z.~Fu, et~al., Deepseek llm: Scaling open-source language models with longtermism.
\newblock arXiv preprint arXiv:2401.02954  (2024)

\bibitem{liu2024deepseek}
A.~Liu, B.~Feng, B.~Xue, B.~Wang, B.~Wu, C.~Lu, C.~Zhao, C.~Deng, C.~Zhang, C.~Ruan, et~al., Deepseek-v3 technical report.
\newblock arXiv preprint arXiv:2412.19437  (2024)

\bibitem{zhang2018personalizing}
S.~Zhang, E.~Dinan, J.~Urbanek, A.~Szlam, D.~Kiela, J.~Weston, \emph{Personalizing Dialogue Agents: I have a dog, do you have pets too?}, in \emph{Proceedings of the 56th Annual Meeting of the Association for Computational Linguistics (Volume 1: Long Papers)} (2018), pp. 2204--2213

\end{thebibliography}

\newpage
\begin{appendices}

\section{Prompts}\label{secA}
We present here the prompts used in our experiments.

\subsection{Topic Generation}\label{secA1}

To generate a diverse conversation topic for each simulation, we used the following prompt:

\noindent\hrulefill
\begin{PromptVerbatim}
Randomly generate a diverse chat topic and give me the topic itself without anything else.
\end{PromptVerbatim}
\noindent\hrulefill

\subsection{Social role generation}\label{secA2}

To create participants with realistic and varied social identities, we used the following prompt:

\noindent\hrulefill
\begin{PromptVerbatim}
Generate a specific social identity and social role that is ordinary, common, diverse and follows real social distribution patterns. The name should be 'Person {i+1}'
(Important: use this name directly, do not create another real name!)
Respond with only the description, no additional text or markdown.
\end{PromptVerbatim}
\noindent\hrulefill

\subsection{Utterance generation with chat history}\label{secA3}
Given a social role, topic, and prior chat history, we used the following prompt:

\noindent\hrulefill
\begin{PromptVerbatim}
You are assigned the following role:
Person {i}
{Social Role}

The chat topic is:
{Topic}

Here is the previous conversation history:
{Chat History}

Based on your role, you have two options:
1. Contribute to the conversation.
2. Remain silent and listen.

If your contribution would be repetitive, irrelevant, or if the topic seems concluded, respond only with: "Stay silent".  
Otherwise, respond with: "Speak: [your message]"

Guidelines:
- Keep the response brief (2-3 sentences).  
- Do not introduce yourself or restate your role.  
\end{PromptVerbatim}
\noindent\hrulefill

\subsection{Utterance generation without chat history}\label{secA4}
For the first utterance of a conversation (no chat history), we used:

\noindent\hrulefill
\begin{PromptVerbatim}
You are assigned the following role:
Person {i}
{Social Role}

The chat topic is:
{Topic}

Respond with: "Speak: [your message]"

Guidelines:
- Keep the response brief (2-3 sentences).  
- Do not introduce yourself or restate your role.  
\end{PromptVerbatim}
\noindent\hrulefill

\subsection{Occupation classification using ISCO-08}\label{secA5}

\noindent\hrulefill
\begin{PromptVerbatim}
You are an expert in occupational classification.
Below is the ISCO-08 classification of occupations into 10 major groups.
Each group has a code, title, and definition.

Your task: Given a description of a social role, choose the single most appropriate group by providing the ISCO code and title.

### ISCO-08 Major Groups ###
Code: 1
Title: Managers
Definition: Managers plan, direct, coordinate and evaluate the overall activities of enterprises, governments and other organizations, or of organizational units within them, and formulate and review their policies, laws, rules and regulations. Competent performance in most occupations in this major group requires skills at the fourth ISCO skill level, except for Sub-major group 14: Hospitality, Retail and Other Services Managers, for which skills at the third ISCO skill level are generally required.
Tasks include: Tasks performed by managers usually include: formulating and advising on the policy, budgets, laws and regulations of enterprises, governments and other organizational units; establishing objectives and standards and formulating and evaluating programmes and policies and procedures for their implementation; ensuring appropriate systems and procedures are developed and implemented to provide budgetary control; authorising material, human and financial resources to implement policies and programmes; monitoring and evaluating performance of the organization or enterprise and of its staff; selecting or approving the selection of  staff; ensuring compliance with health and safety requirements; planning and directing daily operations; representing and negotiating on behalf of the government, enterprise or organizational unit managed in meetings and other forums.
Included occupations: Occupations in this major group are classified into the following sub-major groups:
11 Chief executives, Senior Officials and Legislators
12 Administrative and Commercial Managers
13 Production and Specialized Services Managers
14 Hospitality, Retail and Other Services Managers

Code: 2
Title: Professionals
Definition: Professionals increase the existing stock of knowledge; apply scientific or artistic concepts and theories; teach about the foregoing in a systematic manner; or engage in any combination of these activities. Competent performance in most occupations in this major group requires skills at the fourth ISCO skill level.
Tasks include: Tasks performed by professionals usually include: conducting analysis and  research, and developing concepts, theories and operational methods; advising on or applying existing knowledge related to physical sciences, mathematics, engineering and technology, life sciences, medical and health services, social sciences and humanities; teaching the theory and practice of one or more disciplines at different educational levels; teaching and educating persons with learning difficulties or special needs; providing various business, legal and social services; creating and performing works of art; providing spiritual guidance; preparing scientific papers and reports. Supervision of other workers may be included.
Included occupations: Occupations in this major group are classified into the following sub-major groups:
21 Science and Engineering Professionals
22 Health Professionals
23 Teaching Professionals
24 Business and Administration Professionals
25 Information and Communications Technology Professionals
26 Legal, Social and Cultural Professionals

Code: 3
Title: Technicians and Associate Professionals
Definition: Technicians and associate professionals perform technical and related tasks connected with research and the application of scientific or artistic concepts and operational methods, and government or business regulations. Competent performance in most occupations in this major group requires skills at the third ISCO skill level.
Tasks include: Tasks performed by technicians and associate professionals usually include: undertaking and carrying out technical work connected with research and the application of concepts and operational methods in the fields of physical sciences including engineering and technology, life sciences including the medical profession, and social sciences and humanities; initiating and carrying out various technical services related to trade, finance and administration including administration of government laws and regulations, and to social work; providing technical support for the arts and entertainment; participating in sporting activities; executing some religious tasks. Supervision of other workers may be included.
Included occupations: Occupations in this major group are classified into the following sub-major groups:
31 Science and Engineering Associate Professionals
32 Health Associate Professionals
33 Business and Administration Associate Professionals
34 Legal, Social, Cultural and Related Associate Professionals
35 Information and Communications Technicians

Code: 4
Title: Clerical Support Workers
Definition: Clerical support workers record, organise, store, compute and retrieve information, and perform a number of clerical duties in connection with money-handling operations, travel arrangements, requests for information, and appointments. Competent performance in most occupations in this major group requires skills at the second ISCO skill level.
Tasks include: Tasks performed by clerical support workers usually include: stenography, typing, and operating word processors and other office machines; entering data into computers; carrying out secretarial duties; recording and computing numerical data; keeping records relating to stocks, production and transport; keeping records relating to passenger and freight transport; carrying out clerical duties in libraries; filing documents; carrying out duties in connection with mail services; preparing and checking material for printing; assisting persons who cannot read or write with correspondence; performing money-handling operations; dealing with travel arrangements; supplying information requested by clients and making appointments; operating a telephone switchboard. Supervision of other workers may be included.
Included occupations: Occupations in this major group are classified into the following sub-major groups:
41 General and Keyboard Clerks
42 Customer Services Clerks
43 Numerical and Material Recording Clerks
44 Other Clerical Support Workers

Code: 5
Title: Service and Sales Workers
Definition: Service and sales workers provide personal and protective services related to travel, housekeeping, catering, personal care, or protection against fire and unlawful acts, or demonstrate and sell goods in wholesale or retail shops and similar establishments, as well as at stalls and on markets. Competent performance in most occupations in this major group requires skills at the second ISCO skill level.
Tasks include: Tasks performed by service and sales workers usually include: organizing and providing services during travel; housekeeping; preparing and serving of food and beverages; caring for children; providing personal and basic health care at homes or in institutions, as well as hairdressing, beauty treatment and companionship; telling fortunes; embalming and arranging funerals; providing security services and protecting individuals and property against fire and unlawful acts; enforcing of law and order; posing as models for advertising, artistic creation and display of goods; selling goods in wholesale or retail establishments, as well as at stalls and on markets; demonstrating goods to potential customers. Supervision of other workers may be included.
Included occupations: Occupations in this major group are classified into the following sub-major groups:
51 Personal Service Workers
52 Sales Workers
53 Personal Care Workers
54 Protective Services Workers

Code: 6
Title: Skilled Agricultural, Forestry and Fishery Workers
Definition: Skilled agricultural, forestry and fishery workers grow and harvest field or tree  and shrub crops, gather wild fruits and plants, breed, tend or hunt animals, produce a variety of animal husbandry products; cultivate, conserve and exploit forests; breed or catch fish; and cultivate or gather other forms of aquatic life in order to provide food, shelter and income for themselves and  their households. Competent performance in most occupations in this major group requires skills at the second ISCO skill level.
Tasks include: Tasks performed by skilled agricultural, forestry and fishery workers usually include: preparing the soil; sowing, planting, spraying, fertilizing and harvesting field crops; growing fruit and other tree and shrub crops; growing garden vegetables and horticultural products; gathering wild fruits and plants; breeding, raising, tending or hunting animals mainly to obtain meat, milk, hair, fur, skin, sericultural, apiarian or other products; cultivating, conserving and exploiting forests; breeding or catching fish; cultivating or gathering other forms of aquatic life; storing and carrying out some basic processing of their produce; selling their products to purchasers, marketing organisations or at markets. Supervision of other workers may be included.
Included occupations: Occupations in this major group are classified into the following sub-major groups:
61 Market-oriented Skilled Agricultural Workers
62 Market-oriented Skilled Forestry, Fishery and Hunting Workers
63 Subsistence Farmers, Fishers, Hunters and Gatherers

Code: 7
Title: Craft and Related Trades Workers
Definition: Craft and related trades workers apply specific technical and practical knowledge and skills in the fields to construct and maintain buildings; form metal; erect metal structures; set machine tools or make, fit, maintain and repair machinery, equipment or tools; carry out printing work; and produce or process foodstuffs, textiles and wooden, metal and other articles, including handicraft goods. Competent performance in most occupations in this major group requires skills at the second ISCO skill level.
The work is carried out by hand and by hand-powered and other tools which are used to reduce the amount of physical effort and time required for specific tasks, as well as to improve the quality of the products. The tasks call for an understanding of all stages of the production process, the materials and tools used, and the nature and purpose of the final product.
Tasks include: Tasks performed by craft and related trades workers usually include: constructing, maintaining and repairing buildings and other structures; casting, welding and shaping metal; installing and erecting heavy metal structures, tackle and related equipment; making machinery, tools, equipment and other metal articles; setting for operators, or setting and operating various machine tools; fitting, maintaining and repairing industrial machinery, engines, vehicles, electrical and electronic instruments and other equipment; making precision instruments, jewellery, household and other precious metal articles, pottery, glass and related products; producing handicrafts; executing printing work; producing and processing foodstuffs and various articles made of wood, textiles, leather and related materials. Supervision of other workers may be included.  Self-employed craft and related trades workers, who operate their own businesses either independently or with assistance from a small number of others, may also perform a range of tasks associated with management of the business, account and record keeping and client service, although such tasks would not normally comprise the major component of the work.
Included occupations: Occupations in this major group are classified into the following sub-major groups:
71 Building and Related Trades Workers (excluding Electricians)
72 Metal, Machinery and Related Trades Workers
73 Handicraft and Printing Workers
74 Electrical and Electronic Trades Workers
75 Food processing, Woodworking, Garment and Other Craft and Related Trades Workers

Code: 8
Title: Plant and Machine Operators, and Assemblers
Definition: Plant and machine operators, and assemblers operate and monitor industrial and agricultural machinery and equipment on the spot or by remote control; drive and operate trains, motor vehicles and mobile machinery and equipment; or assemble products from component parts according to strict specifications and procedures. Competent performance in most occupations in this major group requires skills at the second ISCO skill level.

The work mainly calls for experience with and an understanding of industrial and agricultural machinery and equipment as well as an ability to cope with machine-paced operations and to adapt to technological innovations.
Tasks include: Tasks performed by plant and machine operators and assemblers usually include: operating and monitoring mining or other industrial machinery and equipment for processing metal, minerals, glass, ceramics, wood, paper or chemicals; operating and monitoring machinery and equipment used to produce articles made of metal, minerals, chemicals, rubber, plastics, wood, paper, textiles, fur or leather, and which process foodstuffs and related products; driving and operating trains and motor vehicles; driving, operating and monitoring mobile industrial and agricultural machinery and equipment; assembling products from component parts according to strict specifications and procedures. Supervision of other workers may be included.
Included occupations: Occupations in this major group are classified into the following sub-major groups:
81 Stationary Plant and Machine Operators
82 Assemblers
83 Drivers and Mobile Plant Operators

Code: 9
Title: Elementary Occupations
Definition: Elementary occupations involve the performance of simple and routine tasks which may require the use of hand-held tools and considerable physical effort. Most occupations in this major group require skills at the first ISCO skill level.
Tasks include: Tasks performed by workers in elementary occupations usually include: cleaning, restocking supplies and performing basic maintenance in apartments, houses, kitchens, hotels, offices and other buildings; washing cars and windows; helping in kitchens and performing simple tasks in food preparation; delivering messages or goods; carrying luggage and handling baggage and freight; stocking vending machines or reading and emptying meters; collecting and sorting refuse; sweeping streets and similar places; performing various simple farming, fishing, hunting or trapping tasks; performing simple tasks connected with mining, construction and manufacturing including product-sorting; packing and unpacking produce by hand and filling shelves; providing various street services; pedalling or hand-guiding vehicles to transport passengers and goods; driving animal-drawn vehicles or machinery. Supervision of other workers may be included.
Included occupations: Occupations in this major group are classified into the following sub-major groups:
91 Cleaners and Helpers
92 Agricultural, Forestry and Fishery Labourers
93 Labourers in Mining, Construction, Manufacturing and Transport
94 Food Preparation Assistants
95 Street and Related Sales and Service Workers
96 Refuse Workers and Other Elementary Workers

Code: 0
Title: Armed Forces Occupations
Definition: Armed forces occupations include all jobs held by members of the armed forces. Members of the armed forces are those personnel who are currently serving in the armed forces, including auxiliary services, whether on a voluntary or compulsory basis, and who are not free to accept civilian employment and are subject to military discipline. Included are regular members of the army, navy, air force and other military services, as well as conscripts enrolled for military training or other service for a specified period.
Tasks include: nan
Included occupations: Occupations in this major group are classified into the following sub-major groups:
01  Commissioned Armed Forces Officers
02  Non-commissioned Armed Forces Officers
03  Armed Forces Occupations, Other Ranks

### Social Role Description ###
{Social Role}

### Your Answer ###
Provide the ISCO code (0-9) only.
\end{PromptVerbatim}
\noindent\hrulefill

\section{Example of simulated conversation}\label{secB}
We present an example of simulated conversation generated by GPT-O3, with both social roles and the chat topic produced by the model itself. This conversation about reviving extinct species and restoring turtle habitats shows LLMs producing idealized social roles (teacher, firefighter, parish leader, engaged grandparent), anchoring discussion around early ethical framing, and maintaining consistently positive, cooperative, and conflict-free dialogue.

\begin{PromptVerbatim}
Social Roles:
Person 1 is a 34-year-old Black woman living in Atlanta, Georgia. She works as a public elementary school teacher, is married with two children, identifies as Christian, earns a middle-income salary, and volunteers as a youth mentor at her neighborhood community center.

Person 2 is a 29-year-old second-generation Mexican American man living in Chicago, Illinois. He works full-time as an assistant manager at a large retail store, rents an apartment with two roommates, identifies as Catholic, is single but in a long-term relationship, speaks both English and Spanish at home, and volunteers on weekends with a local food pantry.

Person 3 is a 45-year-old Korean American woman living in Seattle, Washington. She works full-time as a registered nurse at a major urban hospital, owns a townhouse with her spouse, has one teenage son, identifies as Buddhist, earns a solid middle-class income, and regularly participates in a local hiking group while also coordinating care for her aging parents who live nearby.

Person 4 is a 52-year-old white man living in a small town outside Dayton, Ohio. He works as a unionized assembly-line technician at an auto parts factory, holds an associate degree from a community college, is married with one adult daughter who attends college, identifies as Protestant, enjoys fishing and watching NFL games, and serves part-time as a volunteer firefighter for his township.

Person 5 is a 63-year-old Puerto Rican woman living in Orlando, Florida. Recently retired after three decades as a medical office receptionist, she receives Social Security and a modest pension, shares a home with her adult son and helps care for her two elementary-aged grandchildren, attends Mass weekly at a local Catholic parish, manages type 2 diabetes, enjoys salsa dancing at community centers, and participates in a neighborhood gardening club that donates produce to a nearby food bank.

Topic:
The ethics and future of resurrecting extinct species through DNA cloning

Conversation History:
Turn 1 - Person 1: Speak: While the idea of bringing back extinct animals is exciting, we need to think hard about whether we're honoring God's creation or risking more imbalance in nature. I'd rather see us preserve the species we still have and restore their habitats so my students can experience a healthy, living planet.

Turn 2 - Person 2: Speak: I agree stewardship is key; before we play with mammoth DNA, we should prove we can safeguard the animals God still entrusts to us. If the research also helps endangered species right now, then it could be worth exploring, but otherwise I'd rather we fix the damage we've already done.

Turn 3 - Person 3: Speak: From a Buddhist sense of interconnectedness, I'm uneasy with reviving animals whose ecosystems no longer exist-it could create more imbalance and suffering. If the cloning techniques directly bolster the health of current endangered species or restore genetic diversity, then tightly regulated pilot projects might be justified. Otherwise our energy and funding should go to protecting the habitats and species we still share the planet with.

Turn 4 - Person 4: Speak: I love the idea of seeing a woolly mammoth, but I'd feel better if the scientists first showed they can keep our local wetlands healthy enough for bass and bluegill. As a firefighter, I know it's smarter to stop a small blaze than try to undo a full-on wildfire, so let's spend the money keeping today's species from going extinct in the first place. If the same research ends up helping living animals survive, great-otherwise it can wait.

Turn 5 - Person 5: Speak: My grandkids light up at the idea of a real-life mammoth, yet Pope Francis keeps reminding us to heal the Earth we already share. If this cloning work directly strengthens today's endangered species-or brings medical insights that help people-then maybe it's worth cautious trials, but let's not fund miracles in a freezer while living creatures and their habitats are still in danger.

Turn 6 - Person 1: Speak: I'm heartened we share the same stewardship focus. If the cloning work can tangibly help protect the gopher frogs and indigo snakes my students study, I'd support limited, transparent trials-but only once we've matched every lab dollar with real investment in restoring the places those living creatures call home.

Turn 7 - Person 2: Speak: Matching lab funds with on-the-ground conservation sounds fair-showing real commitment, not just headline science. If any de-extinction project can prove it's boosting genetic tools to save, say, our Illinois Blanding's turtles right now, I'd back a small pilot. Until then, let's direct most resources to rebuilding habitats and stopping current extinctions first.

Turn 8 - Person 3: Speak: Pairing any de-extinction trials with equivalent funding for habitat restoration feels like the kind of mindful balance the Bodhisattva ethic calls for-alleviating present suffering while exploring future possibilities. If the researchers also commit to long-term monitoring and transparent public reporting, I'd support a small pilot that directly aids endangered species in our own ecosystems.

Turn 9 - Person 4: Speak: Matching every lab dollar with hands-on habitat work feels like the right kind of safety net. If the scientists commit to that-and let regular folks like us keep an eye on the results-I'd be okay with a small, carefully watched pilot.

Turn 10 - Person 5: Speak: Pairing lab money with habitat fixes is a solid start, but let's also make sure any spinoff medical advances-say, insights that could help folks like me manage diabetes-stay affordable and widely shared. If the scientists commit to that and long-term care for the revived animals, I'd bless a careful pilot.

Turn 11 - Person 1: Speak: Absolutely-equitable access to any medical breakthroughs and lifelong care for the animals must be baked into the plan. If the pilot pairs every research dollar with habitat work, transparent public oversight, and guarantees affordable benefits for both wildlife and people, I'm ready to support moving forward cautiously.

Turn 12 - Person 2: Speak: If we add a community advisory panel-bringing in local conservationists, parish leaders, and even school kids-so neighbors can track every dollar and outcome, I'm on board. With that transparency plus the dollar-for-dollar habitat commitment, let's start with a pilot that first boosts Illinois' Blanding's turtles before we dream about woolly mammoths.

Turn 13 - Person 3: Speak: A community advisory panel resonates with the Buddhist principle of collective responsibility; letting students and neighbors witness the impacts firsthand deepens mindfulness and accountability. With that in place alongside the dollar-for-dollar habitat work, I'm comfortable endorsing the pilot focused on Blanding's turtles as a measured next step.

Turn 14 - Person 4: Speak: Count me in. If our township's kids can help track turtle hatchlings and see where every research dollar goes, folks around here will trust the project a whole lot more. Let's get that advisory panel going and make sure the habitat work starts first, then I'm good with the pilot.

Turn 15 - Person 5: Speak: With those safeguards-and a seat for local health clinics and parish reps on the advisory panel so any medical spinoffs truly reach our communities-I'm happy to give the project my blessing. Let's start restoring those turtle wetlands right away so our kids can see hope in action.

Turn 16 - Person 1: Speak: Wonderful-let's set a timeline so the habitat restoration begins this semester, and I'll recruit a few of my fifth-graders to join the advisory panel so they can learn science and stewardship firsthand. If the researchers share clear milestones and budgets with us each quarter, we'll keep everyone accountable and inspired.

Turn 17 - Person 2: Speak: I can rally volunteers from my parish and the pantry crew to pull invasives at the turtle wetlands next month. If we pick a Saturday workday, I'll bring tools and bilingual flyers so everyone feels welcome. Let's lock in a date and I'll handle the sign-ups.

Turn 18 - Person 3: Speak: I can ask my hiking group to join that Saturday-many of us already volunteer on trail crews, so we're comfortable with restoration work. I'll also put together a small first-aid station on site in case anyone needs care. Just let me know the date and I'll coordinate rides and gear.

Turn 19 - Person 4: Speak: I'll check with the township so we can use the firehouse pickup and a portable water tank that day-keeps everyone hydrated and lets us rinse tools. Just give me the Saturday you land on and I'll post the details on our local union board so folks from the plant can pitch in too.

Turn 20 - Person 5: Speak: Just give me the Saturday and I'll come with my son and grandkids-healthy snacks, garden gloves, and a little salsa playlist in tow to keep everyone moving. I can translate for our Spanish-speaking neighbors and have my glucometer handy if anyone needs a quick sugar check. Can't wait to watch those turtle wetlands spring back to life together!

Turn 21 - Person 1: Speak: How about Saturday, April 20th? I'll send permission slips home tomorrow so my fifth-graders can join, and by Friday I'll circulate a contact list and schedule so everyone's set for the workday.

Turn 22 - Person 2: Speak: April 20th is perfect-I'll set up sign-ups at the parish this Sunday and hand out bilingual flyers at the pantry. I'll bring shovels, gloves, and a cooler of water so our crew's ready to work.

Turn 23 - Person 3: Speak: April 20th works for me-I'll post the date to my hiking group tonight and set up carpools so we're there by 9 a.m. I'll also bring a stocked first-aid kit, sunscreen, and an extra pop-up canopy so volunteers have shade while we work.

Turn 24 - Person 4: Speak: Township gave me the green light-the firehouse pickup and 200-gallon water tank will be on site by 8:30 a.m. I'll post sign-ups on the union board first thing tomorrow and bring extra buckets and hoses so we can stay hydrated and keep tools clean. Can't wait to tackle those weeds together.

Turn 25 - Person 5: Speak: April 20th is set-my son and I will be there by 8:45 with fresh fruit, whole-grain snacks, and a salsa playlist to keep us moving. I'll help translate for Spanish speakers and have my glucometer ready in case anyone needs a quick check. Can't wait to see those wetlands and our community thrive side by side!

Turn 26 - Person 1: Speak: Perfect-I'll finalize the permission slips tonight and have my students write thank-you cards for everyone pitching in. I'll arrive by 8:15 to set up a sign-in table and hand out kid-friendly task sheets, so we 're ready to roll when the first truck pulls up.

Turn 27 - Person 2: Speak: Sounds great-I'll get to the site by 8:00 with the flyers, shovels, and water cooler, and I'll text the final volunteer count Friday night so we're ready. I can help your students set up the sign-in table if you'd like.

Turn 28 - Person 3: Speak: I'll be on site by 8:00, set the canopy up next to the sign-in table, and stock the first-aid station with bandages, allergy meds, and extra sunscreen. I'll text everyone Friday with the carpool roster so we know who still needs a ride.

Turn 29 - Person 4: Speak: I'll post the 8 a.m. call time on the union board and remind the night-shift crew they're welcome to stop by on their way home. If anyone's got spare hip waders or weed rakes, toss 'em in the truck-we might need 'em around the soggier patches.

Turn 30 - Person 5: Speak: I'll set up the snack table next to the first-aid canopy and bring a cordless speaker so a little salsa can keep spirits high. Text me any dietary needs by Friday and I'll add options-nos vemos el 20, ready to work and dance!
\end{PromptVerbatim}




\end{appendices}


\end{document}